%% file: main.tex
\documentclass[runningheads]{llncs}
\usepackage[T1]{fontenc}
\usepackage{graphicx}
\usepackage{enumitem}
\usepackage{amsmath}
\usepackage{booktabs}
\usepackage{amssymb}
\usepackage{adjustbox}
\usepackage{bm}
\usepackage{subcaption}
\usepackage{tikz}
\usepackage{xcolor}
\usepackage[hidelinks]{hyperref}
\usepackage{multirow}
\usepackage{xcolor}

\begin{document}
\title{Dual-Foundation Models for Unsupervised Domain Adaptation}

\author{Yerin Cheon\inst{1}\and
Aruna Balasubramanian\inst{1}
\and
Francois Rameau\inst{2}}
\authorrunning{Y. Cheon et al.}

\institute{Stony Brook University, Stony Brook, NY, USA\\ 
\email{\{ycheon, arunab\}@cs.stonybrook.edu}\\
\and
SUNY Korea, The State University of New York, Korea\\
\email{francois.rameau@sunykorea.ac.kr}}
\maketitle             
\input{abstract}
\input{introduction}
\input{related_work}
\input{methods}
\input{experiments}
\input{conclusions}
\input{acknowledgements}

\bibliographystyle{splncs04}
\bibliography{references}
\end{document}

%% file: abstract.tex
\begin{abstract}
Semantic segmentation provides pixel-level scene understanding essential for autonomous driving and fine-grained perception tasks. However, training segmentation models requires costly, labor-intensive annotations on real-world datasets. Unsupervised Domain Adaptation (UDA) addresses this by training models on labeled synthetic data and adapting them to unlabeled real images. While conceptually simple, adaptation is challenging due to the domain gap---differences in visual appearance and scene structure between synthetic and real data.
Prior approaches bridge this gap through pixel-level mixing or feature-level contrastive learning. Yet, these techniques suffer from two major limitations: (1)~reliance on high-confidence pseudo-labels restricts learning to a subset of the target domain, and (2)~prototype-based contrastive methods initialize class prototypes from source-trained models, yielding biased and unstable anchors during adaptation.
To address these issues, we propose a dual-foundation UDA framework that leverages two complementary foundation models. First, we employ the Segment Anything Model (SAM) with superpixel-guided prompting to enable learning from a broader range of target pixels beyond high-confidence predictions. Second, we incorporate DINOv3 to construct stable, domain-invariant class prototypes through its robust representation learning. Our method achieves consistent improvements of +1.3\% and +1.4\% mIoU over strong UDA baselines on GTA$\rightarrow$Cityscapes and SYNTHIA$\rightarrow$Cityscapes, respectively.
Code is available at \url{https://github.com/ycheon1101/DFUDA}.

\keywords{Unsupervised Domain Adaptation  \and Semantic Segmentation \and Foundation Models.}
\end{abstract}

%% file: introduction.tex
\section{Introduction}
\vspace{-0.1cm}
Semantic segmentation refers to the task of assigning a semantic class label to each pixel in an image. It is widely used in scene understanding and robotic perception tasks~\cite{badrinarayanan2016segnetdeepconvolutionalencoderdecoder,mccormac2016semanticfusiondense3dsemantic}.
Training semantic segmentation models typically relies on supervised learning and therefore requires pixel-wise annotations, which are time-consuming and costly to obtain~\cite{cordts2016cityscapes}. In particular, annotating complex real-world scenes containing numerous objects, or scenes captured under adverse weather conditions~\cite{sakaridis2021acdc}, is especially expensive.

To alleviate this issue, labeled synthetic datasets~\cite{richter2016playing,ros2016synthia} are often used to train semantic segmentation models, as they can be generated and annotated at low cost. While models trained on synthetic data typically perform well within the synthetic domain, they fail to generalize directly to real-world data. This lack of generalization is primarily caused by the domain gap between synthetic and real images, which arises from differences in visual style, texture, illumination, and contextual information, as well as from variations between clear and adverse weather conditions~\cite{toldo2020unsupervised}. Unsupervised Domain Adaptation (UDA) has therefore been widely explored to leverage labeled synthetic data while adapting models to unlabeled real-world domains.

To reduce this domain gap, previous works have approached the problem from two major perspectives: pixel-level~\cite{gong2019dlow,kim2020learning,melas2021pixmatch,toldo2020unsupervised,yang2020fda} and feature-level adaptation~\cite{das2023weakly,jiang2022prototypical,paul2020domain,toldo2020unsupervised}.
At the pixel level, many studies employ image mixing strategies~\cite{berthelot2019mixmatch,tranheden2021dacs} that combine source and target images with strong augmentations to mitigate domain differences. However, these approaches typically rely on high-confidence pseudo-labels in the target domain~\cite{tranheden2021dacs}, which restricts learning to a limited subset of target pixels. 
As a result, the model observes only limited target-domain appearance variations that contribute to the domain gap.

At the feature level, contrastive learning~\cite{das2023weakly,jiang2022prototypical,kang2020pixel,wang2021exploring} has been used to align representations between source and target domains. However, pixel-wise contrastive learning~\cite{kang2020pixel,wang2021exploring} requires aligning features for every pixel pair, which is computationally expensive and demands substantial GPU memory. As an alternative, many works adopt prototype-based contrastive learning~\cite{das2023weakly,jiang2022prototypical}.
While more efficient, these methods typically initialize class prototypes from a source-pretrained teacher model~\cite{jiang2022prototypical,zhang2021prototypical}, resulting in source-biased prototypes that do not adequately represent target-domain features. Moreover, prototypes are continuously updated using teacher features during training, making their quality highly dependent on the teacher's predictions. This dependency can lead to prototype collapse when the teacher produces unstable outputs. In addition, prototypes derived from the training network often struggle to represent rare classes, as such categories are infrequently observed during adaptation.

To address these issues, we introduce a framework combining SAM and DINOv3 to strengthen both pseudo-label refinement and feature alignment. SAM provides object-level cues that extend supervision beyond high-confidence target pixels, while DINOv3 offers stable, domain-robust class prototypes that avoid the source bias and instability of teacher-driven prototype updates.
Figure~\ref{fig1} illustrates the overall framework.
Our main contributions are as follows:
\begin{quote}
\begin{itemize}[label=\small\textbullet]
    \item We employ SAM~\cite{kirillov2023segment} to extend supervision beyond high-confidence pseudo-labels using object-level spatial priors.
    \item We introduce a novel mask filtering and superpixel-based prompting strategy to make SAM more efficient and suitable for semantic segmentation.
    \item We construct fixed, domain-agnostic class prototypes from DINOv3~\cite{simeoni2025dinov3} features and use them to guide contrastive alignment, avoiding the source bias and instability of teacher-updated prototypes.
\end{itemize}
\end{quote}

%% file: related_work.tex
\section{Related Work}
\vspace{-0.2cm}
The generalization of semantic segmentation networks across various domains has been a long-lasting problem~\cite{hoyer2022daformerimprovingnetworkarchitectures,hoyer2022hrdacontextawarehighresolutiondomainadaptive,hoyer2023micmaskedimageconsistency,tranheden2021dacs}. 
A common approach to address this issue is unsupervised domain adaptation (UDA), where no labeled data are available in the target domain. In this setting, self-training frameworks have emerged as a popular solution: a teacher network generates pseudo-labels on target images to iteratively train a student model~\cite{bruggemann2023refignalignrefineadaptation,tranheden2021dacs,wang2022continualtesttimedomainadaptation}.
As a result, self-training performance is highly sensitive to the quality of the pseudo-labels. When these labels contain substantial noise, they can reinforce incorrect predictions and lead the model to suffer from confirmation bias~\cite{arpit2017closerlookmemorizationdeep,tarvainen2018meanteachersbetterrole,yang2022divideadaptmitigatingconfirmation}.
In semi-supervised learning and active learning, the use of coarse labels can partially mitigate the performance degradation caused by incorrect pseudo-labels~\cite{das2023weakly}. However, in unsupervised domain adaptation, such coarse labels are not available, making the quality of pseudo-labels even more critical.

Most existing UDA works use the softmax outputs of the model as a confidence measure and train the model using only pseudo-labels whose confidence exceeds a predefined threshold, thereby restricting supervision to high-quality pseudo-labels~\cite{hoyer2022daformerimprovingnetworkarchitectures,10.1007/978-3-030-58568-6_26,tranheden2021dacs,Wang_2020_CVPR}. Some studies additionally employ selective pseudo-labeling strategies~\cite{subhani2020learning,wang2021uncertainty}. Alternatively, prior studies have explored improving generalization performance by refining noisy pseudo-labels using FFT-based techniques~\cite{zhao2023unsuperviseddomainadaptationsemantic} or by explicitly modeling the noise distribution of pseudo-labels~\cite{guo2021metacorrectiondomainawaremetaloss}. Other works increase the model's exposure to rare classes by leveraging class distribution statistics~\cite{hoyer2022daformerimprovingnetworkarchitectures}, thereby enabling more accurate predictions for under-represented classes.

Since the release of SAM~\cite{kirillov2023segment},  it has quickly found use across many tasks to refine model predictions~\cite{lin2025samrefinertamingsegmentmodel,liu2025srplsfdasamguidedreliablepseudolabels,peng2023sam,qin2024langsplat3dlanguagegaussian}.
In UDA, SAM can be employed to refine pseudo-labels using simple majority voting strategies over mask predictions. \cite{sam4udass} further exploits the observation that the relative area ratios of semantic classes remain consistent across domains, and assigns semantic labels to SAM-generated masks based on class-wise area statistics, thereby refining pseudo-labels for the target domain. Additionally, \cite{benigmim2024collaboratingfoundationmodelsdomain} refines pseudo-labels by generating connected-component-based binary masks using the Hoshen–Kopelman algorithm~\cite{hoshen1976percolation}, and then selecting random points within each binary mask as point prompts for SAM. 

Existing SAM-based approaches typically utilize SAM masks merely as binary masks, either to roughly infer semantic classes based on relative area ratios or to determine the presence or absence of objects, thereby primarily focusing on refining object boundaries or improving mask regions. In contrast, our method assigns a unique object-level identifier to each SAM-generated mask, enabling not only boundary refinement but also independent and explicit comparisons across masks during label selection. This design allows for more informed and reliable label assignment at the mask level.

Specifically, in contrast to prior approaches that primarily rely on majority voting or confidence-based heuristics to inject semantic information into SAM, our method adopts a stricter criterion by assuming that each mask corresponds to a single semantic class. By jointly considering softmax probability differences in addition to confidence scores, we favor classes with both low entropy and high confidence, leading to more reliable pseudo-labels. Additionally, we propose a superpixel-guided SAM prompting serves as an efficient mask generation strategy while simultaneously filtering out semantically irrelevant regions, yielding masks that are better suited for semantic segmentation.

In addition to SAM-based pseudo-label refinement, recent works have also explored leveraging self-supervised vision foundation models to mitigate domain gaps. To address domain gaps arising from style and scene-context discrepancies, recent studies have explored leveraging vision foundation models pretrained on large-scale datasets to introduce more robust and generalizable representations~\cite{abedi2024eudaefficientunsuperviseddomain,benigmim2024collaboratingfoundationmodelsdomain,englert2024exploringbenefitsvisionfoundation,fahes2023podapromptdrivenzeroshotdomain,liu2025langdabuildingcontextawarenesslanguage,mata2025coptunsuperviseddomainadaptive,Sikdar_2025_CVPR,10.1145/3664647.3680582,yang2024unifiedlanguagedrivenzeroshotdomain}. ~\cite{benigmim2024collaboratingfoundationmodelsdomain,fahes2023podapromptdrivenzeroshotdomain,10.1145/3664647.3680582,yang2024unifiedlanguagedrivenzeroshotdomain} adopt a frozen CLIP~\cite{radford2021learningtransferablevisualmodels} encoder as the backbone of the segmentation network to extract rich and robust visual representations. Leveraging such vision foundation models has demonstrated strong effectiveness~\cite{englert2024exploringbenefitsvisionfoundation} in domain generalization~\cite{benigmim2024collaboratingfoundationmodelsdomain}, unsupervised domain adaptation (UDA)~\cite{10.1145/3664647.3680582}, and zero-shot domain adaptation~\cite{fahes2023podapromptdrivenzeroshotdomain,yang2024unifiedlanguagedrivenzeroshotdomain}. Furthermore,~\cite{fahes2024simplerecipelanguageguideddomain,Sikdar_2025_CVPR} enhances adaptation to domain discrepancies by exploiting CLIP's powerful global representations while mitigating its limited localized supervision via fine-tuning. Moreover,~\cite{liu2025langdabuildingcontextawarenesslanguage,mata2025coptunsuperviseddomainadaptive} enhance unsupervised domain adaptation performance by encouraging the class representation distances of the segmentation encoder to be close to those in the latent space derived from a frozen CLIP text encoder. 
Closer to our work,~\cite{abedi2024eudaefficientunsuperviseddomain} uses DINOv2~\cite{oquab2024dinov2learningrobustvisual} as a frozen feature extractor for domain-robust representations and improves ViT efficiency with a bottleneck architecture.

Unlike prior works that replace conventional segmentation backbones with large foundation models, we retain a standard segmentation feature extractor and instead incorporate the domain-invariant knowledge of vision foundation models. Specifically, we perform contrastive learning using class-wise prototypes derived from a DINOv3 feature extractor, guiding the model to learn rich and discriminative representations in an efficient manner. Moreover, our approach not only enables the model to follow these informative representations but also simultaneously addresses two long-standing challenges in contrastive learning: source-biased initialization of prototypes and prototype collapse for rare classes.

%% file: methods.tex
\section{Methods}
\vspace{-0.2cm}
The goal of UDA is to adapt a neural network $f_{\theta}$ trained on a labeled source domain $D_S = \{X_S , Y_S\}$ to an unlabeled target domain $D_T = \{X_T\}$, where $X_S = \{x_S\}_{i=1}^{N_S}$ and $Y_S = \{y_S\}_{i=1}^{N_S}$ denote source images with corresponding one-hot segmentation labels, and $X_T = \{x_T\}_{i=1}^{N_T}$ denotes unlabeled target images.

\begin{figure}[t]
\centering
\includegraphics[width=\textwidth]{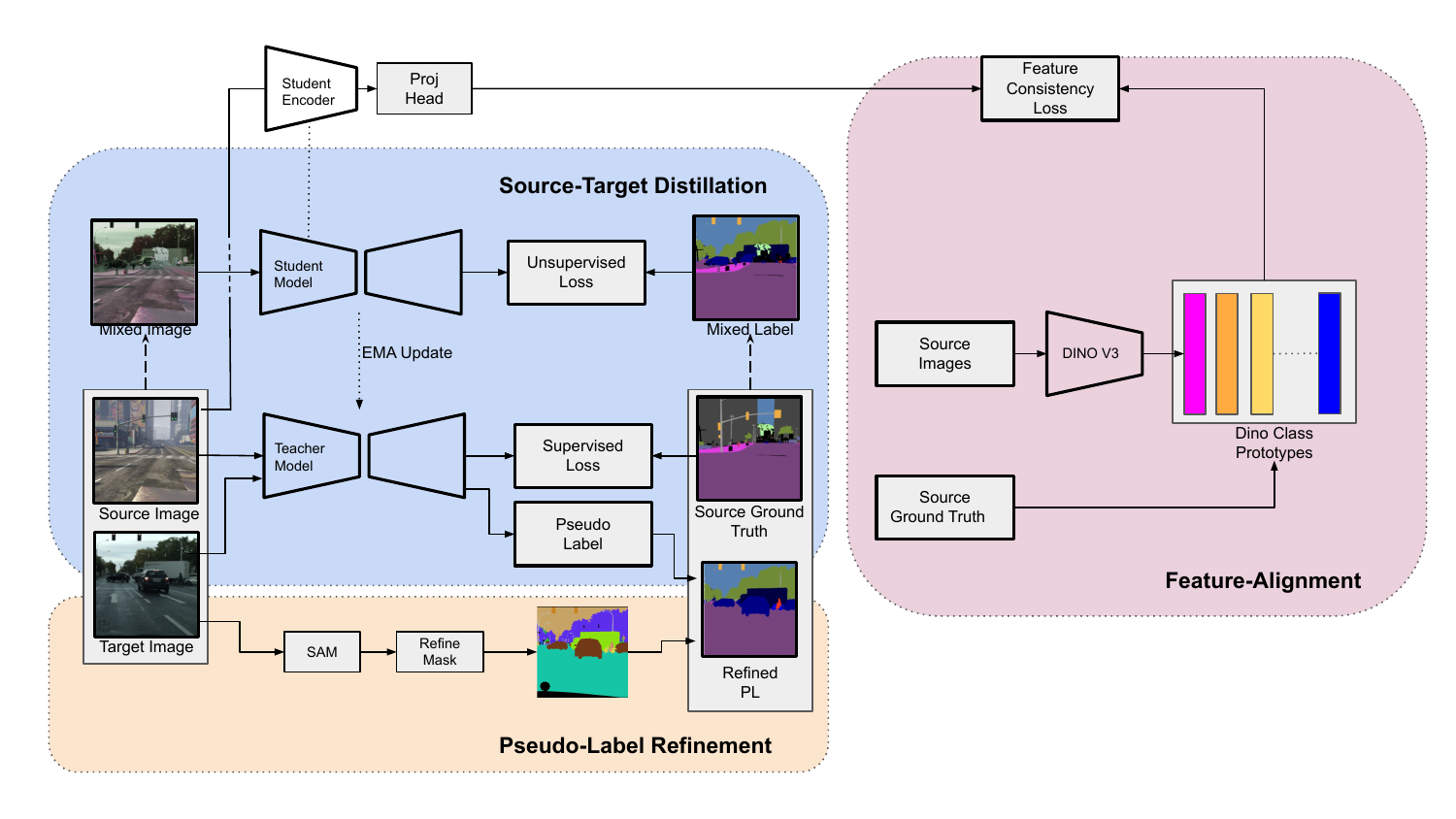}
\caption{\textbf{Overview of our framework.}
(Blue) \textit{Source--Target Distillation:} An online self-training scheme reduces the source--target domain gap using EMA-updated teacher predictions. 
(Yellow) \textit{Pseudo-Label Refinement:} Superpixel-prompted SAM masks are filtered, and each mask region is assigned a high-confidence, low-entropy pseudo-label. (Sec.~\ref{sec::pseudo_label} \& Sec.~\ref{sec::superpixel_sam})
(Pink) \textit{Feature Alignment:} Student features are projected into the same latent space as the DINOv3 prototypes and aligned with DINOv3-based class prototypes to learn domain-invariant representations, reducing source bias and improving generalization. (Sec.~\ref{sec::feature_alignment})
All components interact cyclically, progressively improving pseudo-label quality and adaptation performance. 
} \label{fig1}
\end{figure}

Following recent state-of-the-art methods~\cite{hoyer2022daformerimprovingnetworkarchitectures,hoyer2022hrdacontextawarehighresolutiondomainadaptive,hoyer2023micmaskedimageconsistency,tranheden2021dacs}, we adopt an online self-training strategy to reduce the domain gap. Our framework extends existing approaches through three key components. First, while conventional methods rely solely on confidence-based pseudo-label filtering, we refine teacher-generated pseudo-labels using SAM mask regions. For each mask, we assign pseudo-labels that satisfy both high confidence and low entropy criteria, where entropy is measured by the softmax probability gap (Sec.~\ref{sec::pseudo_label}). Second, we introduce superpixel-guided prompting with overlap-aware mask filtering to efficiently generate semantically coherent masks suitable for segmentation (Sec.~\ref{sec::superpixel_sam}). Third, we leverage class-wise prototypes derived from DINOv3 features to guide the model toward learning stable, domain-invariant representations (Sec.~\ref{sec::feature_alignment}). \figurename~\ref{fig1} illustrates our complete pipeline.

\vspace{-0.35cm}

\subsection{Source--Target Distillation with Self-Training} 
We build upon the standard teacher--student self-training  framework~\cite{das2023weakly,hoyer2022daformerimprovingnetworkarchitectures,hoyer2022hrdacontextawarehighresolutiondomainadaptive,hoyer2023micmaskedimageconsistency,tranheden2021dacs}.  A student network $f_{\theta}$ is trained using supervised loss $\mathcal{L}_S$ on labeled source images and unsupervised loss $\mathcal{L}_T$ on target images pseudo-labeled by a teacher network $h_{\phi}$. For the target loss computation, we select only pixels where the teacher's prediction confidence exceeds a threshold $\tau$. The teacher parameters $\phi$ are updated via exponential moving average (EMA) of the student parameters $\theta$, maintaining stable pseudo-label generation throughout training. In the following sections, we describe how we enhance this baseline with SAM-based pseudo-label refinement and DINOv3-guided feature alignment.

\vspace{-0.1cm}
\subsection{Pseudo-Label Refinement Using SAM} 
\label{sec::pseudo_label}
To overcome the limitation of prior methods that train models only on high-quality target pixels, thereby exposing the student to only a small portion of the target domain, we leverage SAM masks generated by our proposed approach (Sec.~\ref{sec::superpixel_sam}). Specifically, in this section, we introduce our approach for refining pseudo-labels by assigning appropriate semantic labels to SAM masks, which do not natively carry any semantic information. 

Each pixel $(h',w')$ is assigned a mask ID, where SAM masks are indexed from $1$ to $M$ (the number of masks) and uncovered pixels are assigned ID $0$:

\begin{equation}
\text{MaskID}^{(h',w')} =
\begin{cases}
m, & \text{if pixel } (h',w') \in m, \\
0, & \text{otherwise}.
\end{cases}
\label{eq:mask_id}
\end{equation}

\noindent For every mask region, our goal is to assign a single semantic class label. To ensure that each mask receives a reliable and consistent label, we select only the pixels satisfying two confidence criteria.

\paragraph{(1) High Softmax Confidence.}
For each pixel $(h',w')$ in a mask region, we first check whether its maximum softmax probability exceeds a threshold $\tau$.
\begin{equation}
C_1^{(h',w')} = 
\left[\, \max_{c'} \, h_{\phi}(x_T)^{(h',w',c')} > \tau \, \right].
\label{eq:conf_thresh}
\end{equation}
\paragraph{(2) Low Entropy via Softmax Margin.}
Inspired by~\cite{Kweon_2024_CVPR}, we sort the softmax probabilities at pixel $(h',w')$ 
and compute the difference between the highest and second-highest probabilities. We denote the largest and second-largest probabilities at pixel $(h', w')$ as $p_{(1)}^{(h', w')}$ and $p_{(2)}^{(h', w')}$, respectively (i.e., the top-1 and top-2 entries of $\mathbf{p}^{(h', w')}$).
Only pixels with a gap above threshold $\tau'$ are retained.

\begin{equation}
C_2^{(h',w')} = \left[\, p_{(1)}^{(h',w')} - p_{(2)}^{(h',w')} > \tau' \, \right].
\label{eq:entropy_thresh}
\end{equation}

\noindent For each mask region, we collect pixels satisfying both confidence conditions in Eq.~\ref{eq:conf_thresh} and Eq.~\ref{eq:entropy_thresh}. For each selected pixel $(h'',w'')$, the pseudo-label is defined as:
\begin{equation}
PL^{(h'',w'', c)} = \left[\ c =  \arg\max_{c'} \, h_{\phi}(x_T)^{(h'',w'',c')} \right].
\label{eq:pl_selected}
\end{equation}

\noindent We then check whether all selected pixels within the same mask share the same pseudo-label. 
If all selected pixels agree on the same class $k$, the entire mask region is assigned that class. Otherwise, we discard the mask refinement and preserve the original pixel-level pseudo-labels.

\begin{figure}[t]
\centering

\begin{subfigure}[t]{0.3\textwidth}
\centering
\includegraphics[width=\linewidth]{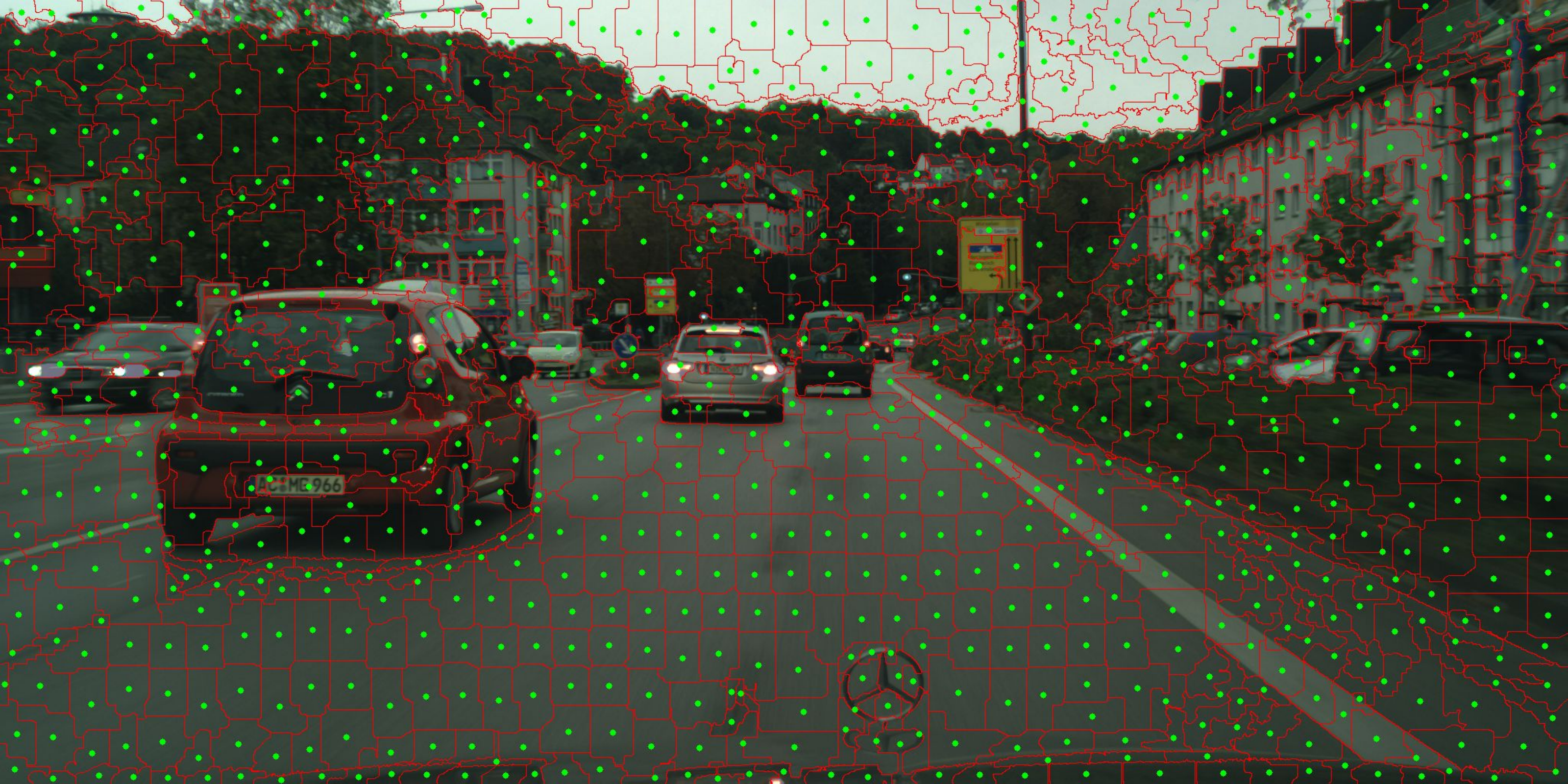}
\caption{}
\label{fig:superpixel_prompts}
\end{subfigure}
\begin{subfigure}[t]{0.3\textwidth}
\centering
\includegraphics[width=\linewidth]{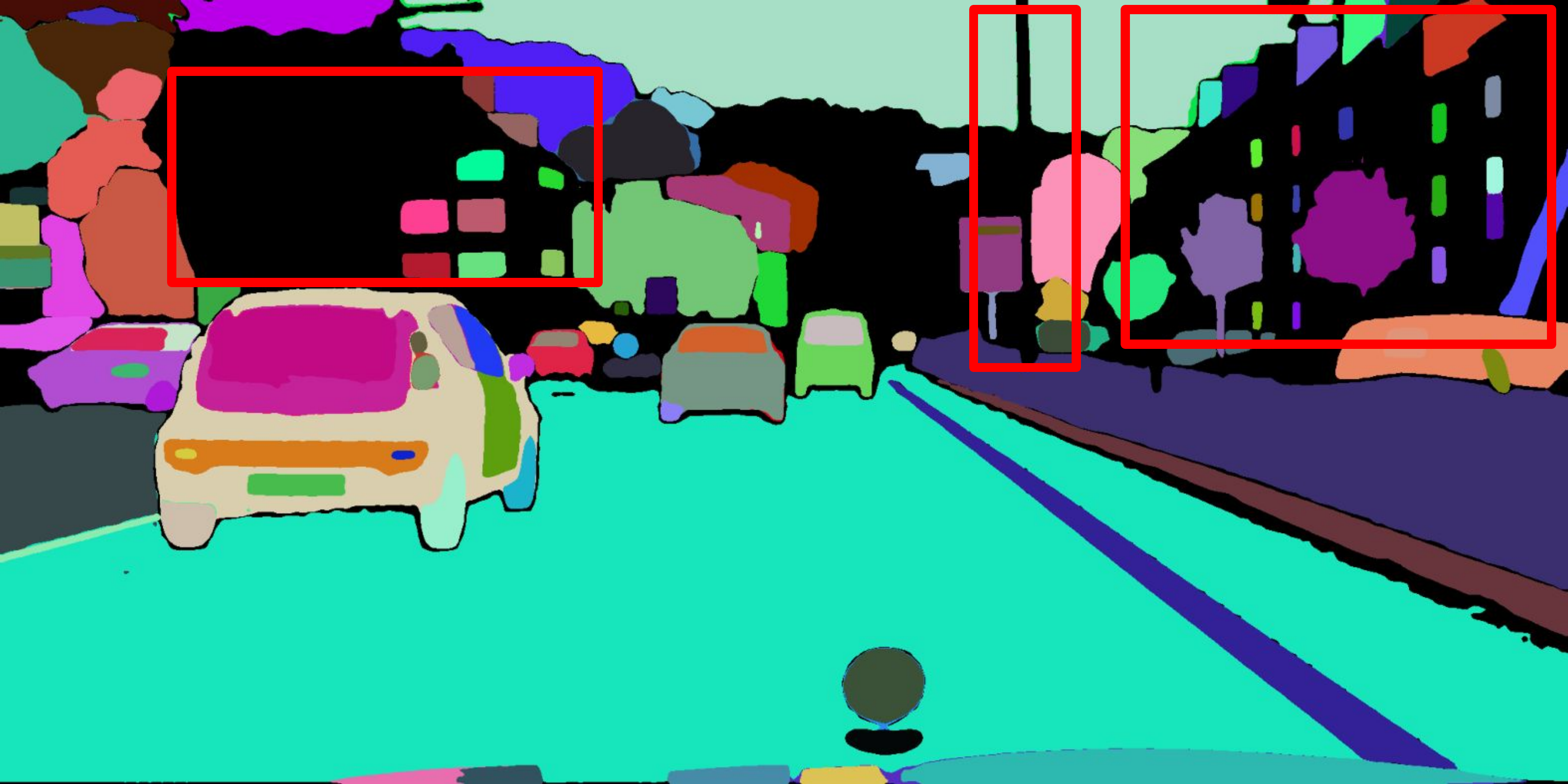}
\caption{}
\label{fig:sam_auto_prompt}
\end{subfigure}
\begin{subfigure}[t]{0.3\textwidth}
\centering
\includegraphics[width=\linewidth]{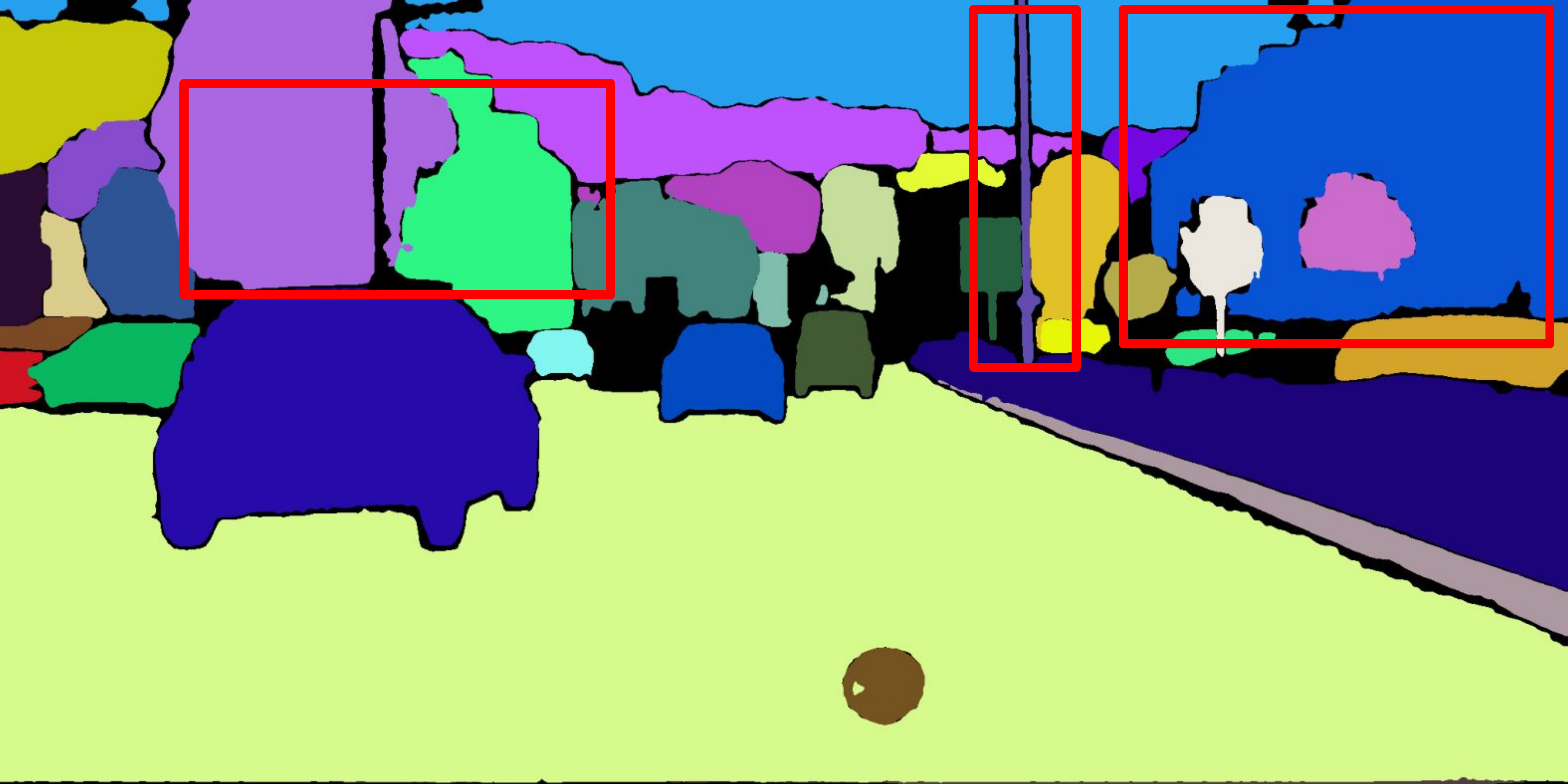}
\caption{}
\label{fig:sam_superpixel}
\end{subfigure}

\caption{\textbf{Comparison of SAM mask generation strategies.}
(a) Superpixel-based SAM prompting (Sec.~\ref{sec::superpixel_sam}).
(b) SAM automatic mask generation.
(c) Our superpixel-guided and filtered masks.
The proposed method yields more meaningful and structurally coherent masks than the SAM auto mask generator.}
\label{fig:superpixel_and_sam}
\vspace{-0.5cm}
\end{figure}

\subsection{Super-Pixel based SAM and Filtering Mask} 
\vspace{-0.25cm}
\label{sec::superpixel_sam}
To address the limitations of SAM in semantic segmentation, where it tends to produce overly fragmented and redundant masks, we propose Superpixel-guided SAM, which integrates a superpixel-based point sampling strategy into the automatic mask generation process, followed by overlap-aware filtering and region refinement. Our approach generates a compact and structurally coherent set of masks that is better suited for downstream semantic segmentation.\\
\\
{\bfseries Superpixel-guided Point Prompts.}
\\
Given an input image $I \in \mathbb{R}^{H \times W \times 3}$, we partition the image into $K$ superpixels using the SEEDS~\cite{bergh2013seedssuperpixelsextractedenergydriven} algorithm: $\mathcal{S} = \{ S_1, S_2, \dots, S_K \}.$
Each superpixel $S_k$ is a spatially contiguous set of pixels. We define a representative center point for each superpixel as $\mathbf{p}_k = \left(\operatorname{median}_{(x,y)\in S_k}(x), \operatorname{median}_{(x,y)\in S_k}(y)
\right)$.
To match the normalized point coordinate format required by SAM, the center points are normalized to $\tilde{\mathbf{p}}_k = (x_k / W,\; y_k / H)$.
The resulting superpixel-guided point set is defined as: $\mathcal{P} = \{ \tilde{\mathbf{p}}_k \}_{k=1}^{K}$.
By using these points as prompts, we generate a more meaningful and structurally coherent set of masks compared to the standard automatic mask generator (as illustrated in Fig.~\ref{fig:superpixel_and_sam}).
\\
\\
{\bfseries Overlap-aware Mask Filtering and Mask ID Assignment.}
\\
Since SAM tends to generate multiple overlapping masks for the same region, we apply an overlap-aware filtering strategy that retains only non-overlapping regions after area-based sorting. Given a set of candidate masks is $\mathcal{M} = \{ M_i \mid i = 1, \dots, N \}$, 
we first sort the masks in descending order of their area $\mathcal{M}^{\downarrow} = \operatorname{sort}(\mathcal{M}, |M_i|)$.
We maintain a set of already assigned pixels, denoted by $\Omega$, which is initially empty.
For each mask $M_i \in \mathcal{M}^{\downarrow}$,
we compute its non-overlapping region as $\tilde{M}_i = M_i \setminus \Omega$.
If $\lvert \tilde{M}_i \rvert > 0$, the mask is retained,
and the assigned pixel set is updated as $\Omega \leftarrow \Omega \cup \tilde{M}_i$.
This greedy strategy preserves larger regions while naturally removing smaller and redundant masks.

We define an indicator function for each retained mask $\tilde{M}_k$ as
\begin{equation}
\mathbb{I}_{\tilde{M}_k}(x,y) =
\begin{cases}
1, & \text{if } (x,y) \in \tilde{M}_k, \\
0, & \text{otherwise}.
\end{cases}
\end{equation}

\noindent The final label map $L$ is then defined as
\begin{equation}
L(x,y) = \sum_{k=1}^{K} k \cdot \mathbb{I}_{\tilde{M}_k}(x,y),
\end{equation}
where mask IDs are assigned sequentially starting from $k=1$. Pixels that do not belong to any mask satisfy $L(x,y)=0$.

\subsection{Feature Alignment with Domain-Agnostic Prototype}
\label{sec::feature_alignment}
{\bfseries DINO-based Domain-Agnostic Prototype Construction.}
\\
Given a source dataset with semantic labels, we extract dense feature maps using a pretrained DINO encoder $\bm{f} = F_{\text{DINO}}(\mathbf{I}) \in \mathbb{R}^{C \times H' \times W'}$, where $C$ denotes the feature dimension.
For each semantic class $c \in \{1, \dots, K\}$, we compute a class prototype by averaging DINO features belonging to that class:
\begin{equation}
\mathbf{p}_c
=
\frac{1}{|\Omega_c|}
\sum_{(x,y)\in\Omega_c}
\bm{f}_{x,y},
\end{equation}
where $\Omega_c = \{(x,y) \mid y_{x,y} = c\}$
denotes the set of pixels labeled as class $c$. Each prototype vector is subsequently $\ell_2$-normalized to ensure a consistent scale across classes,
i.e., $\mathbf{p}_c \leftarrow \mathbf{p}_c / \lVert \mathbf{p}_c \rVert_2$.
The resulting prototype matrix is then formed as $\boldsymbol{P}
=
[\mathbf{p}_1, \mathbf{p}_2, \dots, \mathbf{p}_K]
\in
\mathbb{R}^{C \times K}$. These prototypes are fixed during training and shared across domains,
serving as domain-agnostic semantic anchors.
\\

\noindent{\bfseries Prototype-based Contrastive Alignment.}\\
For each pixel, we obtain a projected feature vector $\mathbf{z}_i$ by feeding the student encoder feature into a projection head.
We then compute the temperature-scaled cosine similarity
$s_{i,c} = \mathbf{z}_i^\top \mathbf{p}_c / T$.
Based on these similarities, we define the prototype contrastive loss as:
\begin{equation}
\mathcal{L}_{\text{proto}}
=
-\frac{1}{N}
\sum_{i=1}^{N}
\log
\frac{
\exp(s_{i,y_i})
}{
\sum_{c=1}^{K}
\exp(s_{i,c})
}.
\end{equation}
This loss encourages pixel features to be close to their corresponding class prototypes while being separated from prototypes of other classes. The final loss is given by
\begin{equation}
\mathcal{L}
=
\mathcal{L}_S
+
\mathcal{L}_T
+
\lambda \mathcal{L}_{\text{proto}},
\end{equation}
where $\lambda$ is empirically set to $0.1$.

%% file: experiments.tex
\section{Experiments}

\begin{figure}[t]
\centering

\begin{subfigure}[t]{0.22\textwidth}
\centering
\includegraphics[width=\linewidth]{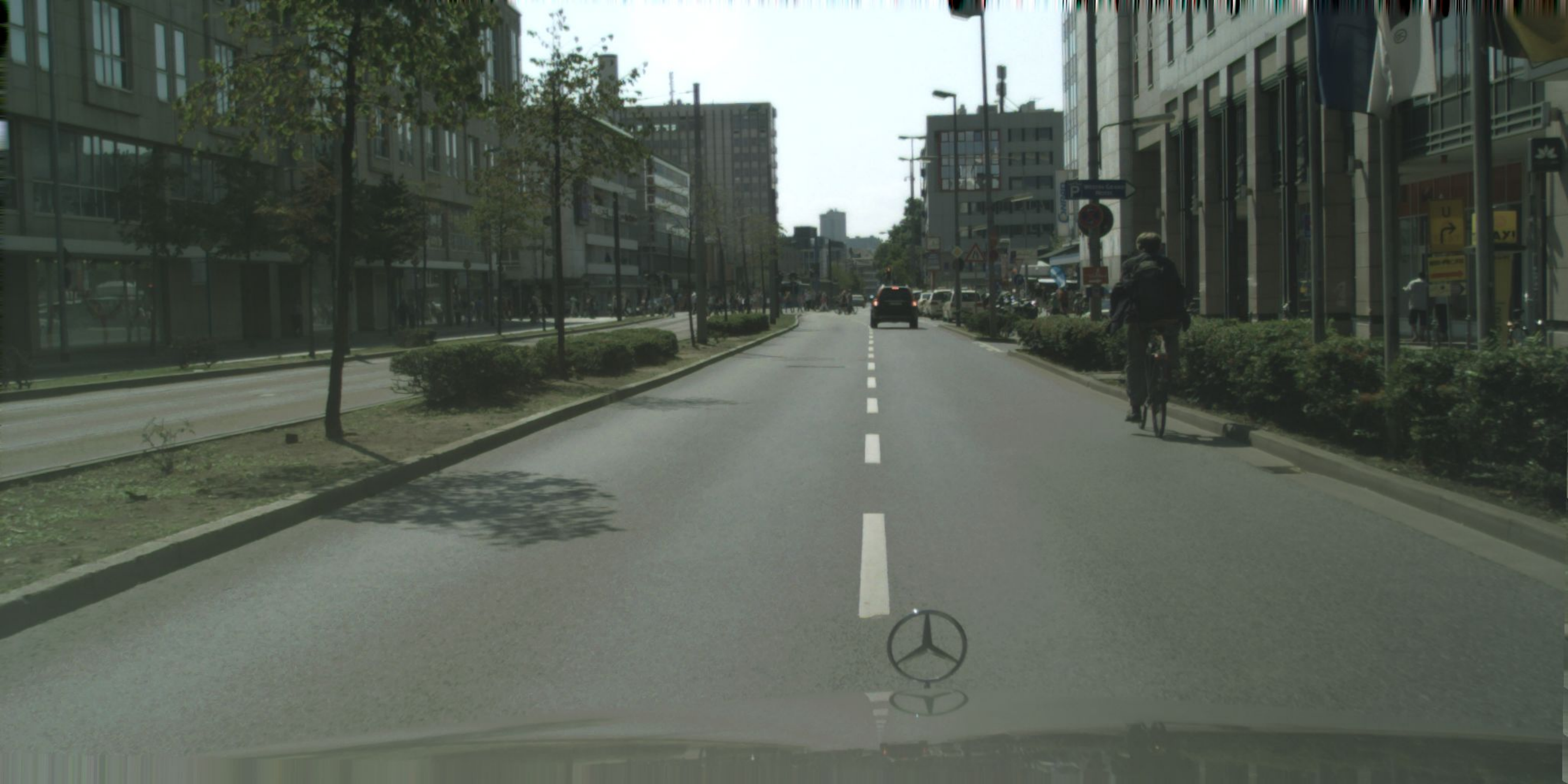}
\caption{Image}
\label{fig:comp_img}
\end{subfigure}
\begin{subfigure}[t]{0.22\textwidth}
\centering
\includegraphics[width=\linewidth]{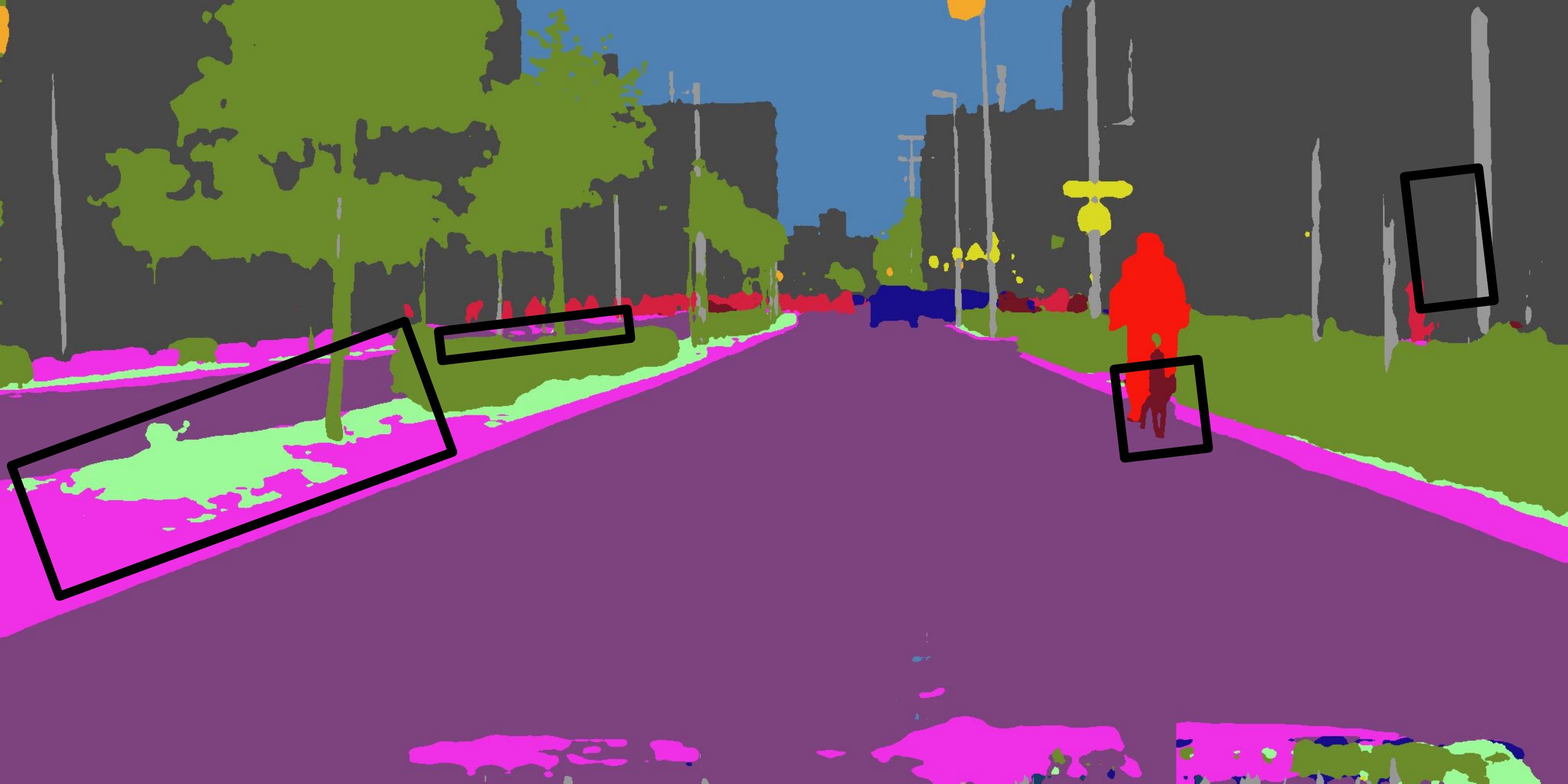}
\caption{MIC~\cite{hoyer2023micmaskedimageconsistency}}
\label{fig:comp_mic}
\end{subfigure}
\begin{subfigure}[t]{0.22\textwidth}
\centering
\includegraphics[width=\linewidth]{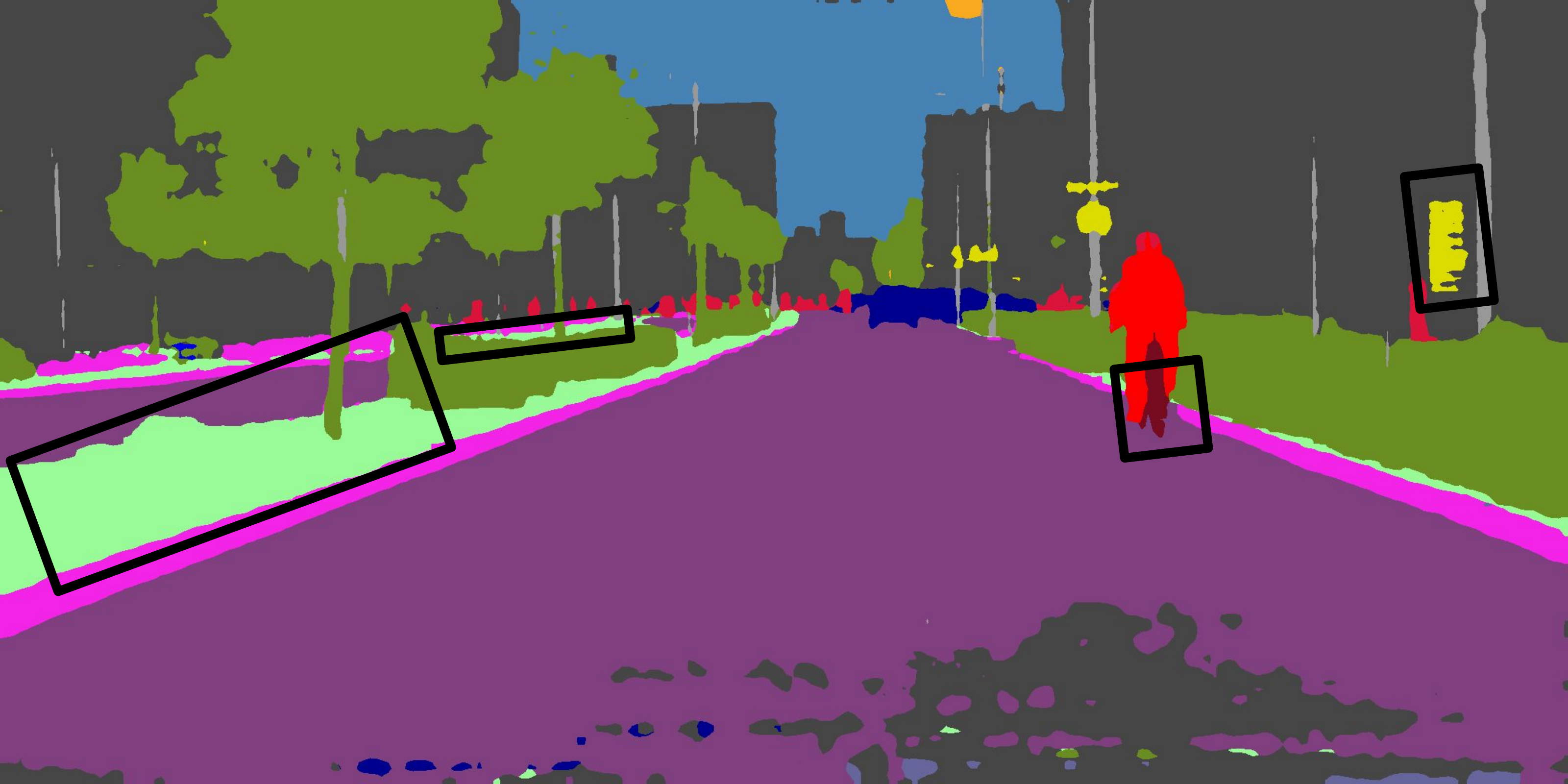}
\caption{Ours}
\label{fig:comp_ours}
\end{subfigure}
\begin{subfigure}[t]{0.22\textwidth}
\centering
\includegraphics[width=\linewidth]{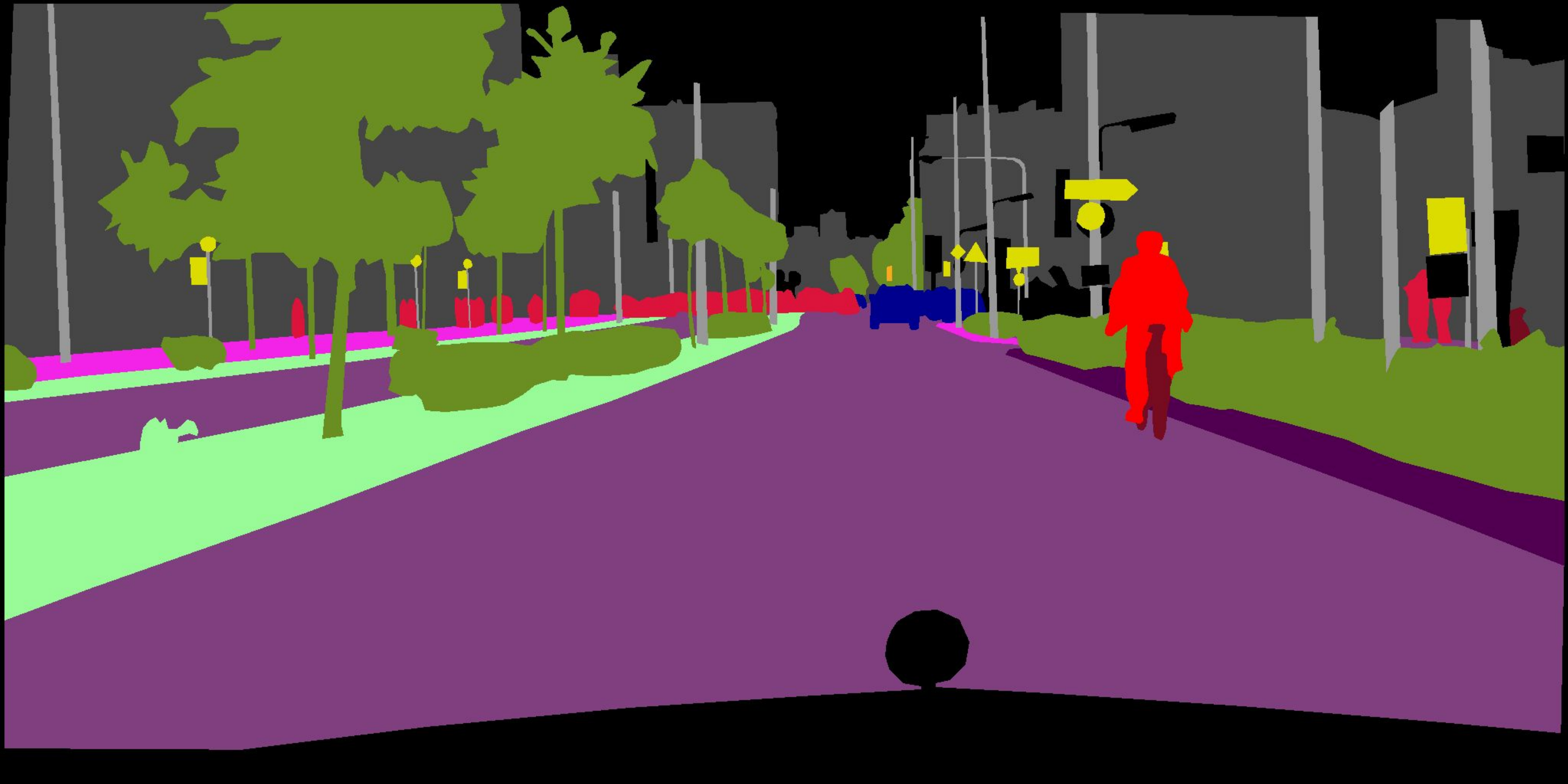}
\caption{Ground Truth}
\label{fig:comp_gt}
\end{subfigure}

\caption{\textbf{Comparison of our method with the state-of-the-art baseline.} Compared to MIC, our method produces improved segmentation for challenging classes such as traffic sign and terrain.
In addition, influenced by SAM-based Pseudo Label refinement, fine structures such as bicycle wheels are more completely filled, closely matching the ground truth.
}
\label{fig:comp_with_mic
}

\end{figure}

\begin{figure}
\centering
\begin{subfigure}[t]{0.14\textwidth}
\centering
\includegraphics[width=\linewidth]{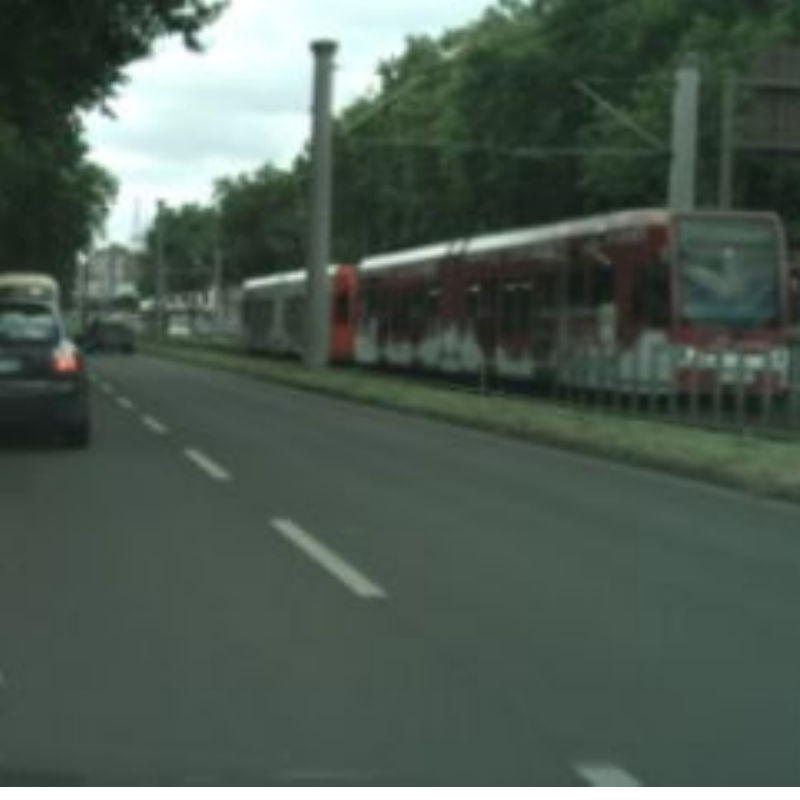}
\label{fig:ab_img1}
\end{subfigure}
\begin{subfigure}[t]{0.14\textwidth}
\centering
\includegraphics[width=\linewidth]{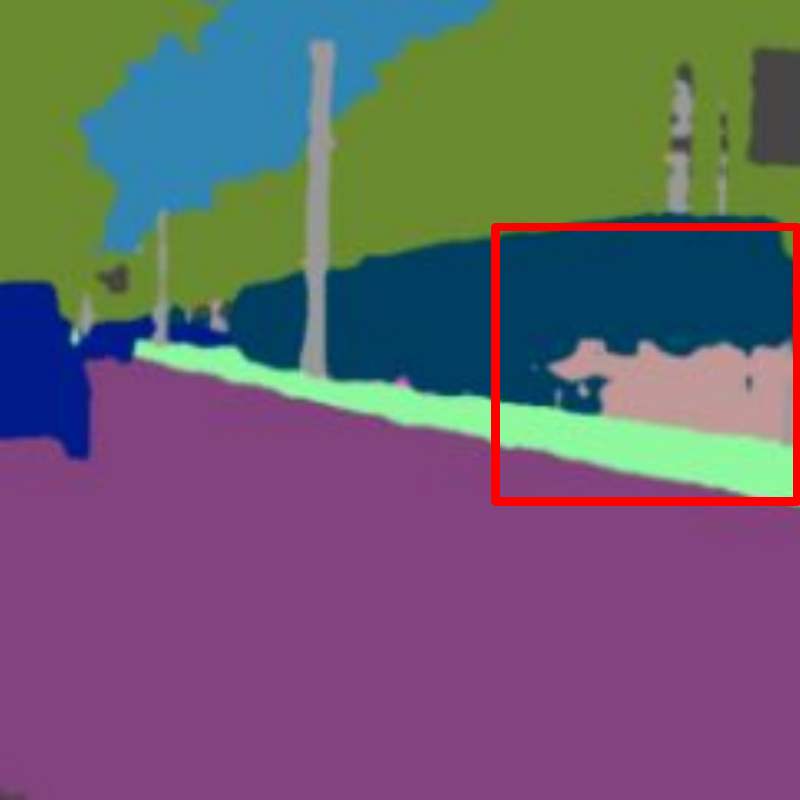}
\label{fig:ab_img1_wo_dinosam}
\end{subfigure}
\begin{subfigure}[t]{0.14\textwidth}
\centering
\includegraphics[width=\linewidth]{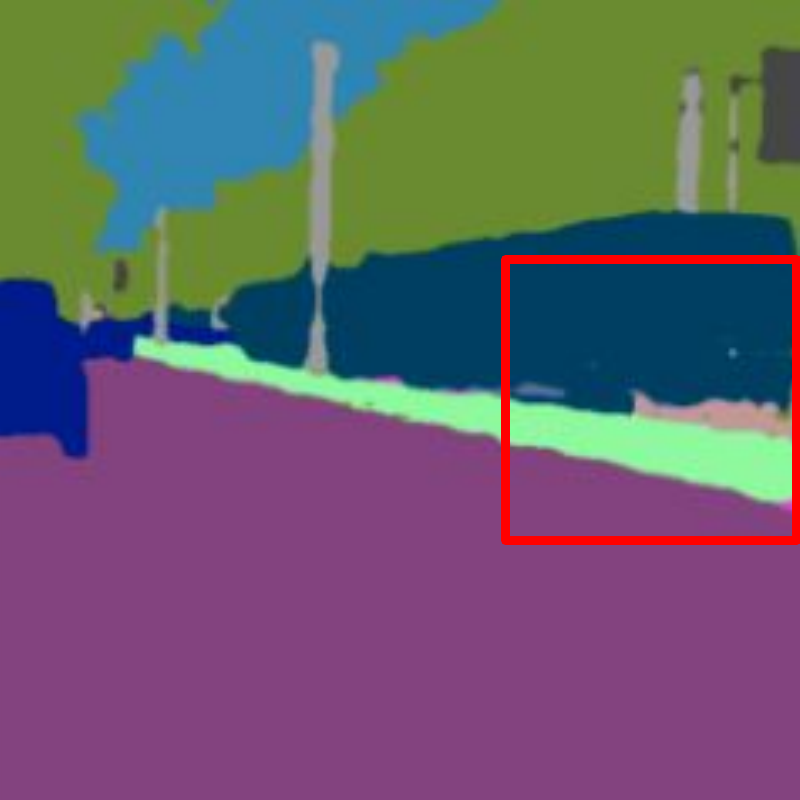}
\label{fig:ab_img1_w_dino}
\end{subfigure}
\begin{subfigure}[t]{0.14\textwidth}
\centering
\includegraphics[width=\linewidth]{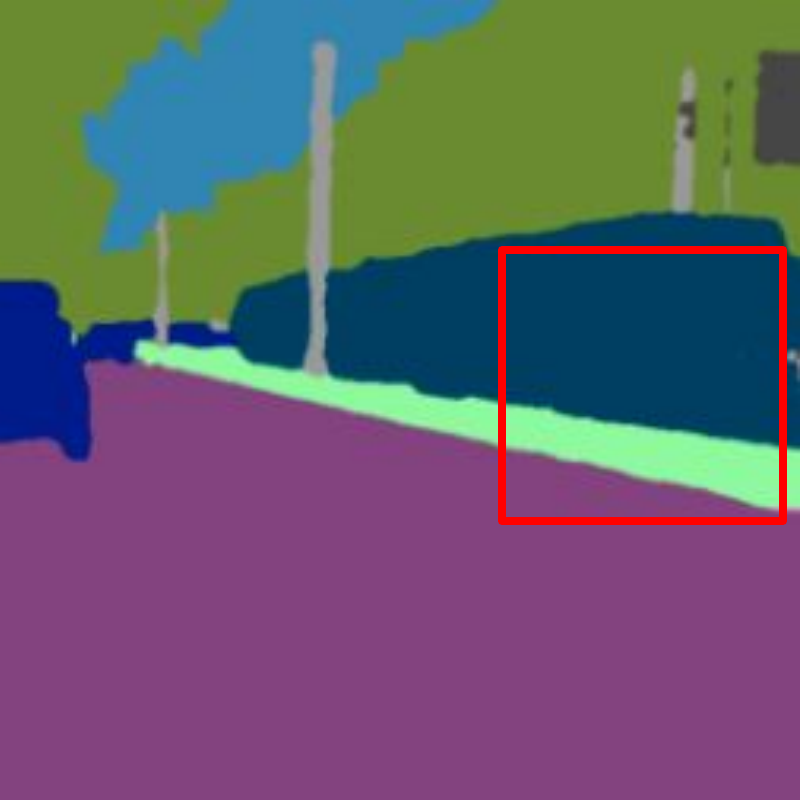}
\label{fig:ab_img1_w_dinosam}
\end{subfigure}

\vspace{-1em}
\begin{subfigure}[t]{0.14\textwidth}
\centering
\includegraphics[width=\linewidth]{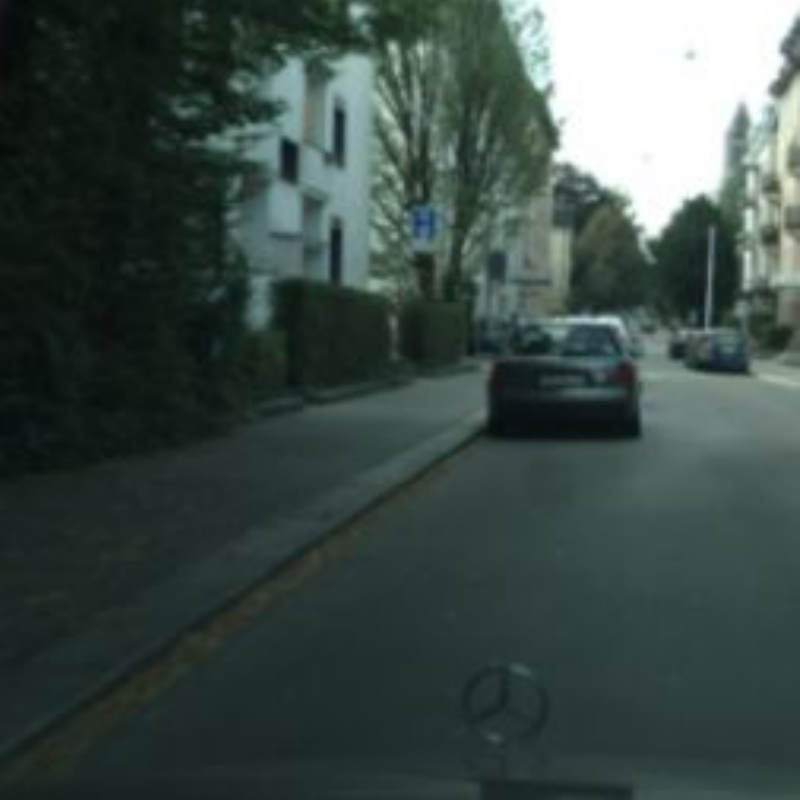}
\caption{}
\label{fig:ab_img2}
\end{subfigure}
\begin{subfigure}[t]{0.14\textwidth}
\centering
\includegraphics[width=\linewidth]{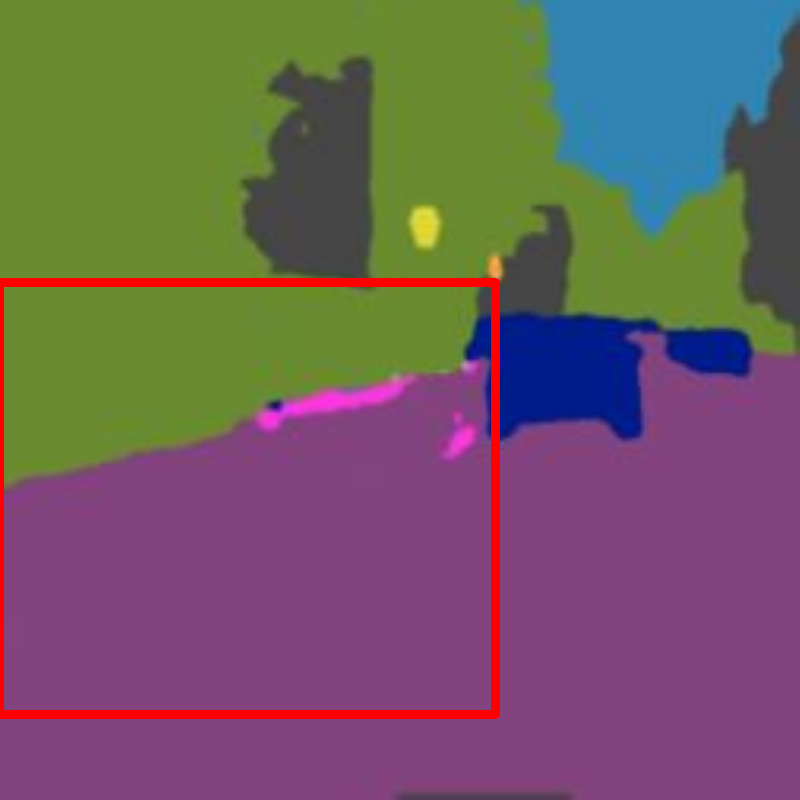}
\caption{}
\label{fig:ab_img2_wo_samdino}
\end{subfigure}
\begin{subfigure}[t]{0.14\textwidth}
\centering
\includegraphics[width=\linewidth]{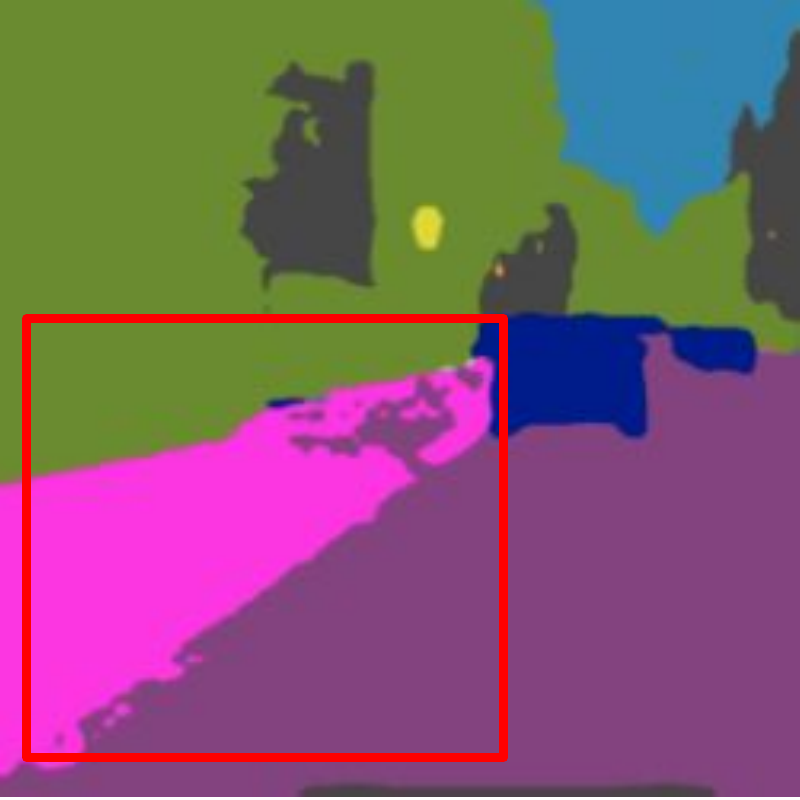}
\caption{}
\label{fig:ab_img2_w_dino}
\end{subfigure}
\begin{subfigure}[t]{0.14\textwidth}
\centering
\includegraphics[width=\linewidth]{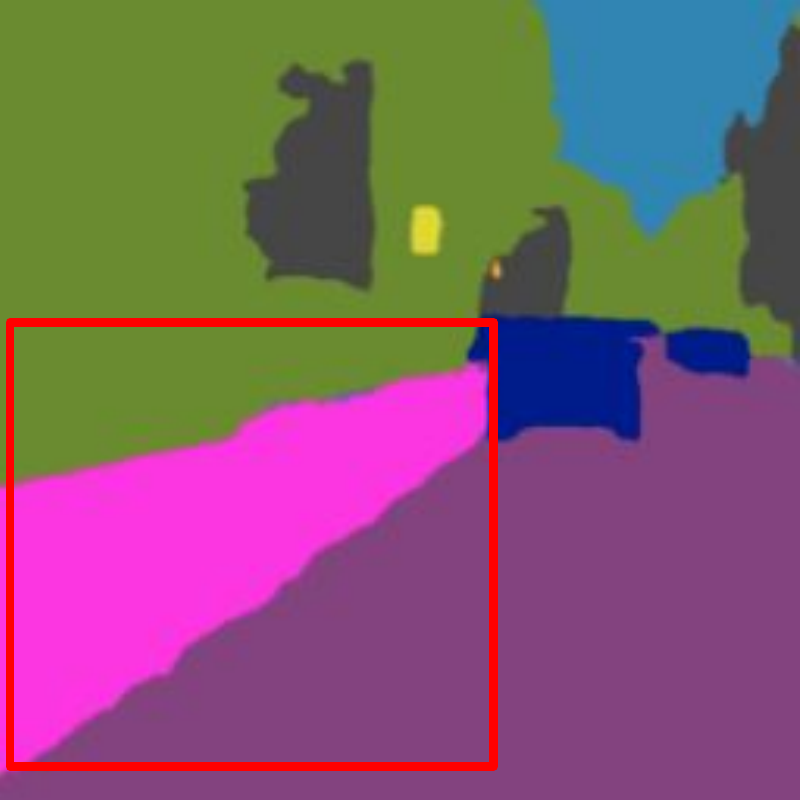}
\caption{}
\label{fig:ab_img2_w_dinosam}
\end{subfigure}
\caption{\textbf{Ablation of DINO-based prototype alignment and Superpixel-based SAM.} (a) Original image.
(b) Without SAM and DINO.
(c) Only DINO-based prototype alignment.
(d) With both SAM and DINO.
DINO improves segmentation in hard-to-distinguish regions via feature alignment, while SAM enhances boundary-aware object prediction.
} \label{fig2}
\vspace{-0.1cm}
\end{figure}
\vspace{-0.3cm}

\subsection{Datasets \& Implementation Details}
We evaluate our method on the Cityscapes dataset~\cite{cordts2016cityscapes} as the target domain, which contains 2,975 training images and 500 validation images with 19 semantic classes.
As source domains, we use GTA with 24,966 images covering all 19 classes, and SYNTHIA~\cite{ros2016synthia} with 9,400 images covering 16 classes.
Our framework is instantiated on two representative UDA architectures: 
DAFormer~\cite{hoyer2022daformerimprovingnetworkarchitectures} with a MiT-B5 encoder, 
and DeepLabV2~\cite{chen2017deeplabsemanticimagesegmentation} with a ResNet-101 backbone, 
both pretrained on ImageNet.
Our base method is MIC~\cite{hoyer2023micmaskedimageconsistency}, but our strategy is applicable on any other self-training-based UDA pipelines (Sec.~\ref{ablation}). The model is optimized using AdamW with learning rates of $6 \times 10^{-5}$ and $6 \times 10^{-4}$ for the encoder and decoder, respectively, and a batch size of 2.
We follow the training protocol of DACS~\cite{tranheden2021dacs}, including data augmentation, EMA with $\alpha=0.99$, and confidence-based pseudo-labeling with threshold $\tau=0.968$.
In addition, we introduce a refined threshold $\tau'=0.99$ in our framework.
Superpixels are generated with a maximum of 1,000 regions per image.
For feature alignment, we extract representations using DINOv3~\cite{simeoni2025dinov3} with a ViT-L backbone, and employ a lightweight $1 \times 1$ convolution as the projection head.
\vspace{-0.2cm}

\subsection{Super-Pixel based SAM}
\begin{figure}
\vspace{-0.3cm}
\centering

\begin{subfigure}[t]{0.3\textwidth}
\centering
\includegraphics[width=\linewidth]{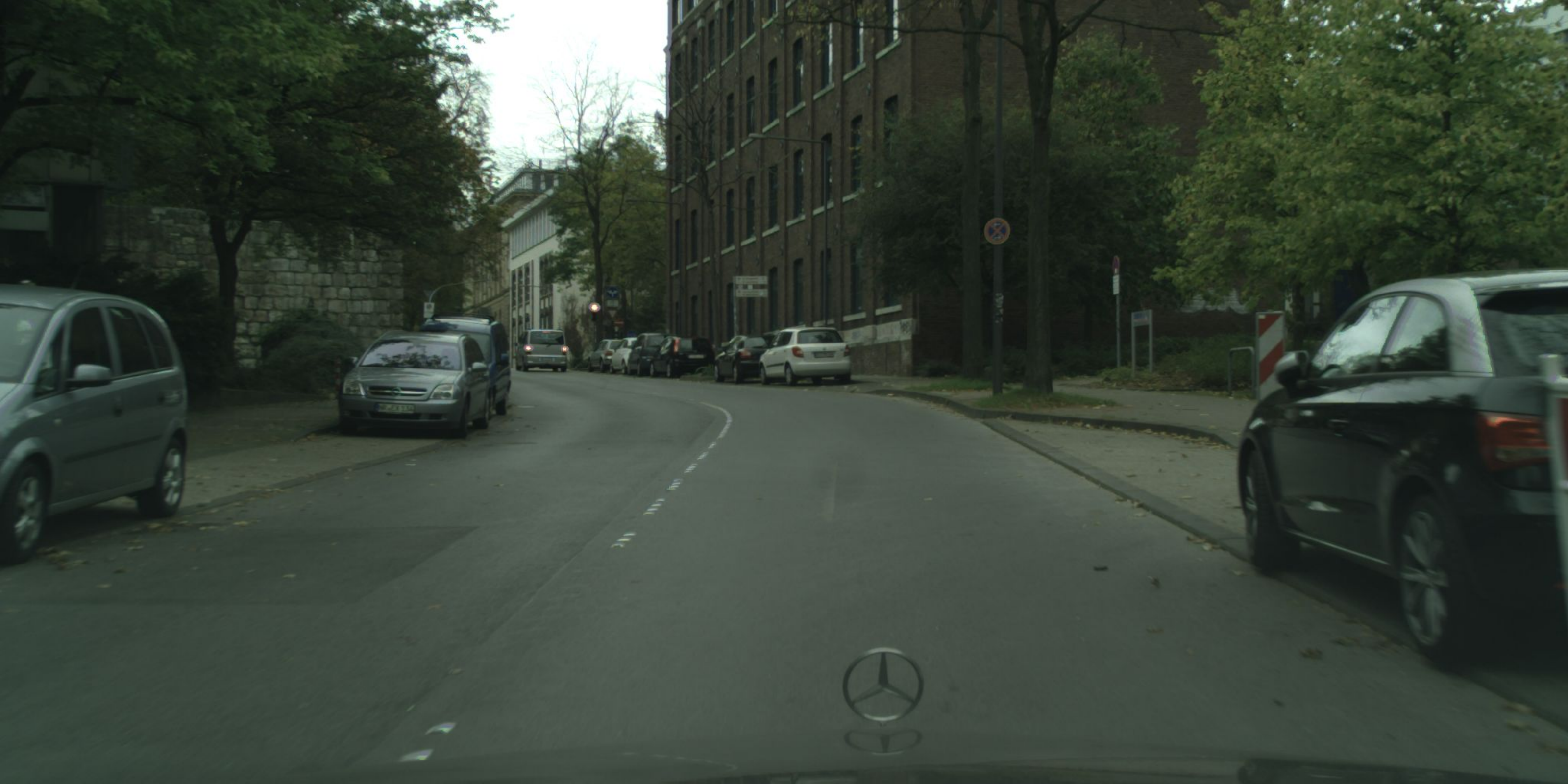}
\label{fig:sam_auto1}
\end{subfigure}
\begin{subfigure}[t]{0.3\textwidth}
\centering
\includegraphics[width=\linewidth]{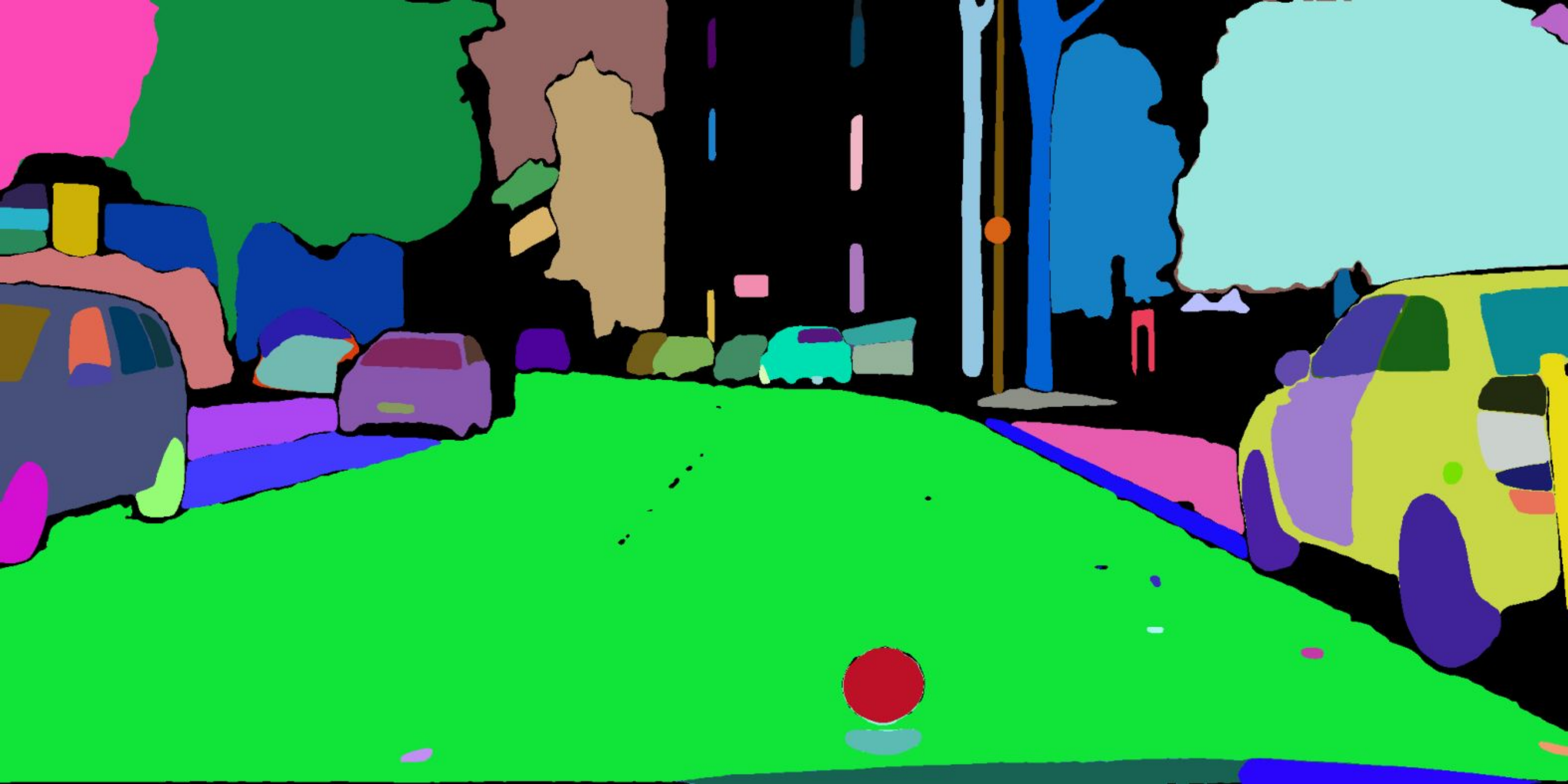}
\label{fig:sam_auto2}
\end{subfigure}
\begin{subfigure}[t]{0.3\textwidth}
\centering
\includegraphics[width=\linewidth]{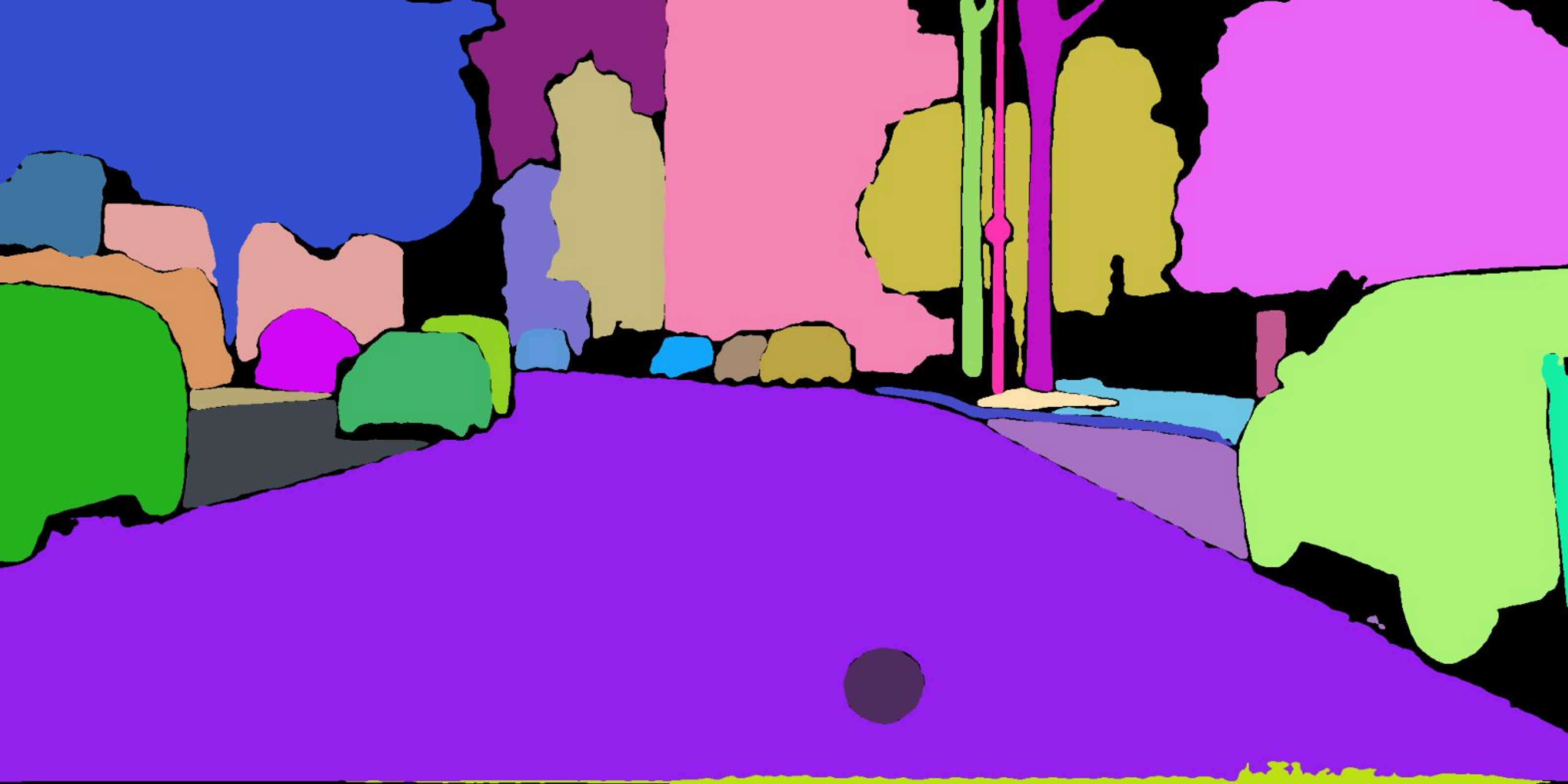}
\label{fig:sam_auto3}
\end{subfigure}

\begin{subfigure}[t]{0.3\textwidth}
\centering
\includegraphics[width=\linewidth]{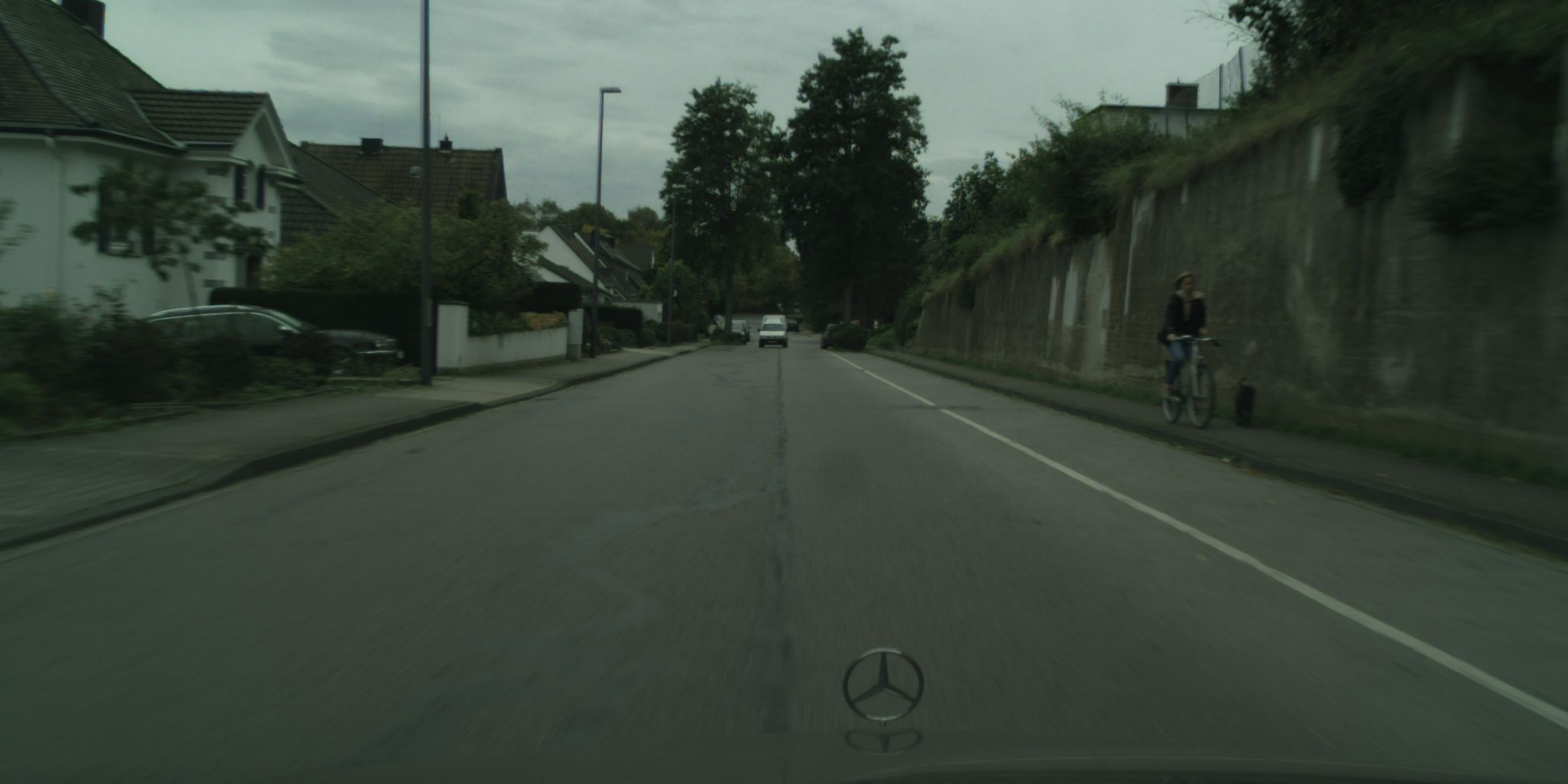}
\caption{Original Image}
\label{fig:sam_refine1}
\end{subfigure}
\begin{subfigure}[t]{0.3\textwidth}
\centering
\includegraphics[width=\linewidth]{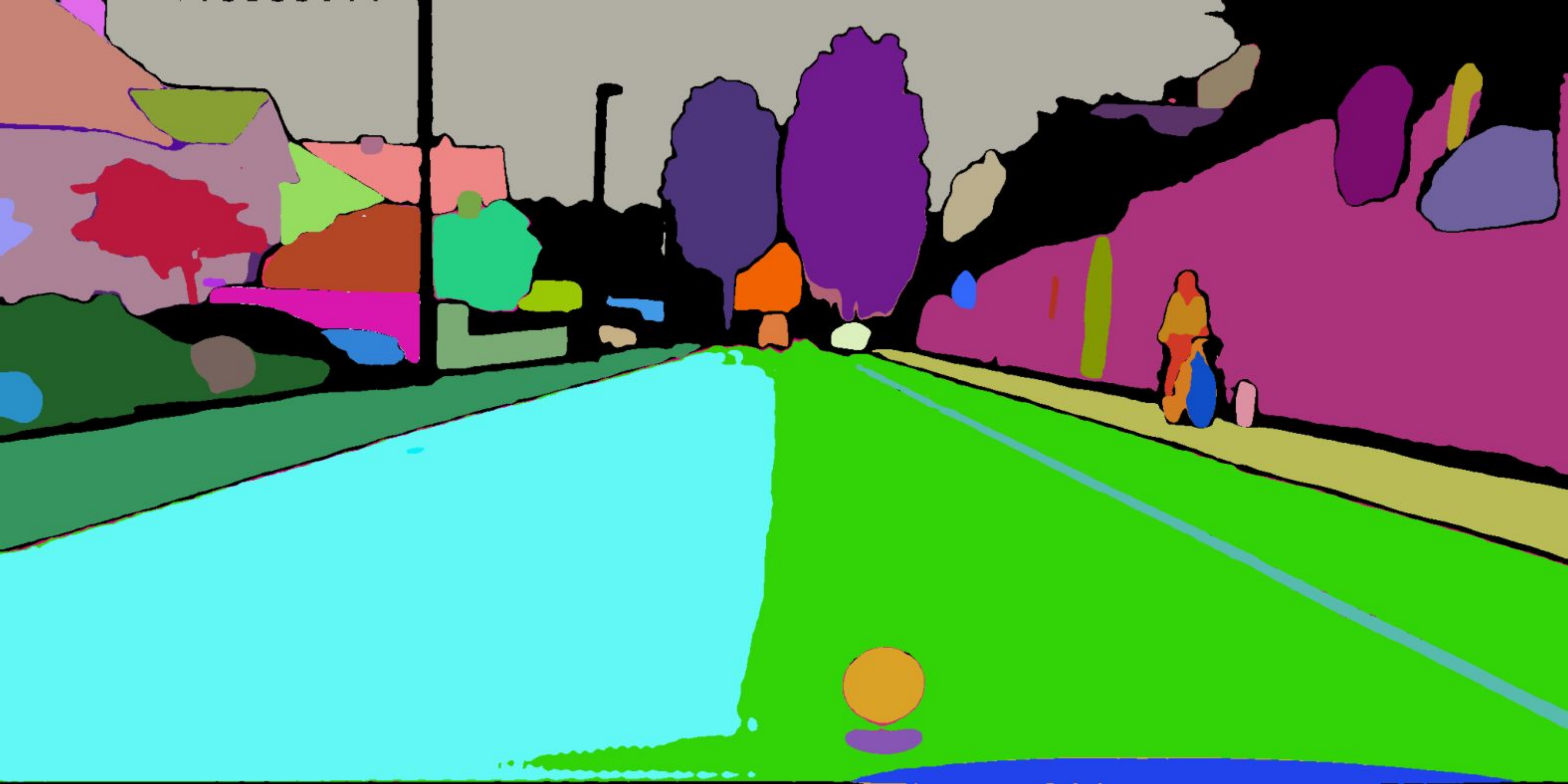}
\caption{SAM AutoMask}
\label{fig:sam_refine2}
\end{subfigure}
\begin{subfigure}[t]{0.3\textwidth}
\centering
\includegraphics[width=\linewidth]{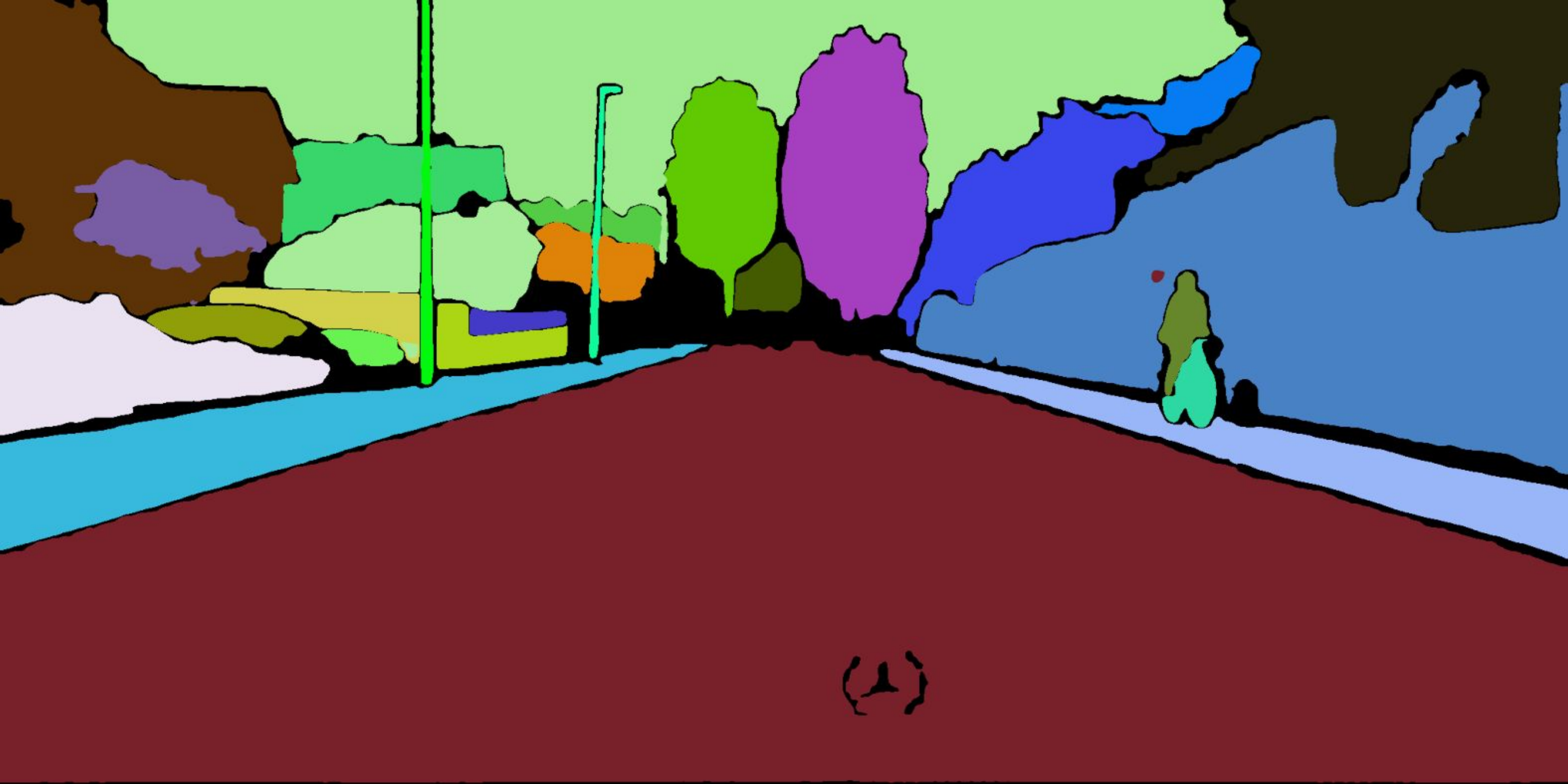}
\caption{Ours}
\label{fig:sam_refine3}
\end{subfigure}

\caption{\textbf{SAM Mask Comparison.}
Comparison between SAM AutoMask and our superpixel-based SAM with overlap-aware filtering.}
\label{fig3}

\end{figure}

In this section, we evaluate the effectiveness of our superpixel-based SAM by comparing it with the SAM automatic mask generator. As shown in Figures~\ref{fig:sam_auto_prompt} and~\ref{fig:sam_refine2}, the automatic mask generator produces overly fine-grained and fragmented masks, which are less suitable for semantic segmentation. Figure~\ref{fig:superpixel_and_sam} qualitatively compares our superpixel-based SAM with filtering (Fig.~\ref{fig:sam_superpixel}) against the SAM automatic mask generator (Fig.~\ref{fig:sam_auto_prompt}). 
As summarized in Table~\ref{tab:mask_quality}, our method uses 13.9\% fewer point prompts than the automatic mask generator while producing a comparable number of final masks. Despite using fewer prompts, our method produces more semantically meaningful object-level masks. The automatic mask generator often segments fine structures such as windows instead of entire buildings and fails to capture thin objects like poles (red boxes in Fig.~\ref{fig:sam_auto_prompt}). In contrast, our approach successfully segments complete objects while avoiding unnecessary over-segmentation.
Table~\ref{tab:mask_quality} further shows that our method improves average mask coverage by +28.52 percentage points, reduces offline mask generation time by 13.4\%, and improves online training throughput by 15.2\% on the Cityscapes training set. These results suggest that superpixel-guided prompting produces more usable object-level masks while reducing the computational overhead of SAM-based refinement. It also remains robust in challenging cases such as closely attached rider--bicycle instances (Fig.~\ref{fig:sam_refine3}).

\vspace{-0.3cm}

\begin{table}[t]
\centering
\scriptsize
\setlength{\tabcolsep}{2pt}
\caption{Comparison of SAM prompting strategies on the Cityscapes train set. Offline mask generation and online training speed are measured on an RTX 4070.}
\label{tab:mask_quality}

\begin{subtable}[t]{0.48\linewidth}
\centering
\caption{Effective mask generation}
\label{tab:effective_mask_generation}
\begin{tabular}{lccc}
\toprule
\multirow{2}{*}{Method} & \# of & \# of & \multirow{2}{*}{Cov.} \\
                        & Prompts & Masks & \\
\midrule
AutoMask & 1024 & 118 & 62.94 \% \\
Ours & 882 & 117 & 91.46 \% \\
\bottomrule
\end{tabular}
\end{subtable}
\hfill
\begin{subtable}[t]{0.48\linewidth}
\centering
\caption{Efficiency}
\label{tab:mask_efficiency}
\begin{tabular}{lcc}
\toprule
\multirow{2}{*}{Method} & \multirow{2}{*}{Mask Gen.} & Train \\
                        &                            & Throughput \\
\midrule
AutoMask & 8.04 s/img & 0.46 it/s \\
Ours & 6.96 s/img & 0.53 it/s \\
\bottomrule
\end{tabular}
\end{subtable}

\end{table}

\vspace{-0.4cm}

\subsection{Unsupervised Domain Adaptation for Semantic Segmentation}
Table~\ref{tab:sota_segmentation} compares our method with state-of-the-art UDA approaches on the GTA$\rightarrow$Cityscapes and SYNTHIA$\rightarrow$Cityscapes benchmarks, where all evaluations are conducted on the Cityscapes validation set using IoU as the evaluation metric.
When applied on top of MIC, our method consistently improves performance across most semantic classes, achieving gains of $+1.3\%$ and $+1.4\%$ mIoU on GTA$\rightarrow$Cityscapes and SYNTHIA$\rightarrow$Cityscapes, respectively.

On GTA$\rightarrow$Cityscapes, our method shows notable improvements on rare or challenging classes such as \emph{train} and \emph{bike}, as well as visually ambiguous classes including \emph{fence}, \emph{pole}, \emph{traffic light}, \emph{traffic sign}, and \emph{sidewalk}.
These improvements indicate that aligning pixel-level features with DINOv3-based prototypes helps the model learn more discriminative and robust representations, particularly for classes that are underrepresented or difficult to distinguish. This is also reflected in the t-SNE visualization in Fig.~\ref{fig:tsne}, which shows improved clustering of target-domain features around DINOv3 prototype anchors. In addition, refining pseudo-labels with SAM enables the model to leverage a larger number of reliable pixels, especially for rare classes.
The combination of SAM-based pseudo-label refinement and DINO-based feature alignment proves particularly effective for these challenging categories. Similar trends are observed on the SYNTHIA$\rightarrow$Cityscapes benchmark.
While many existing UDA methods struggle on the \emph{fence} class, our method achieves a $+2.3\%$ improvement over COPT, suggesting that the proposed framework better transfers the generalization capability of DINO features to the target domain.

\begin{table}
\vspace{-0.3cm}
\centering
\caption{Semantic segmentation performance (IoU in \%) on different UDA benchmarks. All methods' results are taken from publications.}
\label{tab:sota_segmentation}
\setlength{\tabcolsep}{1pt}
\normalsize
\begin{adjustbox}{width=\textwidth}
\begin{tabular}{l|ccccccccccccccccccc|c}
\hline
\toprule
Method & Road & S.walk & Build. & Wall & Fence & Pole & Tr.Light & Sign & Veget. & Terrain & Sky & Person & Rider & Car & Truck & Bus & Train & M.bike & Bike & mIoU\\
\toprule
\multicolumn{21}{c}{\textbf{Synthetic-to-Real: GTA$\to$Cityscapes (Val.)}} \\
\hline
ADVENT~\cite{vu2019adventadversarialentropyminimization} & 89.4 & 33.1 & 81.0 & 26.6 & 26.8 & 27.2 & 33.5 & 24.7 & 83.9 & 36.7 & 78.8 & 58.7 & 30.5 & 84.8 & 38.5 & 44.5 & 1.7 & 31.6 & 32.4 & 45.5\\
DACS~\cite{tranheden2021dacs} & 89.9 & 39.7 & 87.9 & 30.7 & 39.5 & 38.5 & 46.4 & 52.8 & 88.0 & 44.0 & 88.8 & 67.2 & 35.8 & 84.5 & 45.7 & 50.2 & 0.0 & 27.3 & 34.0 & 52.1\\
ProDA~\cite{zhang2021prototypical} & 87.8 & 56.0 & 79.7 & 46.3 & 44.8 & 45.6 & 53.5 & 53.5 & 88.6 & 45.2 & 82.1 & 70.7 & 39.2 & 88.8 & 45.5 & 59.4 & 1.0 & 48.9 & 56.4 & 57.5\\
DAFormer~\cite{hoyer2022daformerimprovingnetworkarchitectures} & 95.7 & 70.2 & 89.4 & 53.5 & 48.1 & 49.6 & 55.8 & 59.4 & 89.9 & 47.9 & 92.5 & 72.2 & 44.7 & 92.3 & 74.5 & 78.2 & 65.1 & 55.9 & 61.8 & 68.3\\
HRDA~\cite{hoyer2022hrdacontextawarehighresolutiondomainadaptive} & 96.4 & 74.4 & 91.0 & 61.6 & 51.5 & 57.1 & 63.9 & 69.3 & 91.3 & 48.4 & 94.2 & 79.0 & 52.9 & 93.9 & 84.1 & 85.7 & 75.9 & 63.9 & 67.5 & 73.8 \\
MIC~\cite{hoyer2023micmaskedimageconsistency} & 97.4 & 80.1 & \underline{91.7} & 61.2 & \underline{56.9} & \underline{59.7} & 66.0 & \underline{71.3} & \underline{91.7} & 51.4 & 94.3 & 79.8 & \textbf{56.1} & 94.6 & 85.4 & \underline{90.3} & 80.4 & 64.5 & 68.5 & 75.9\\
COPT~\cite{mata2025coptunsuperviseddomainadaptive} & \underline{97.6} & \underline{80.9} & 91.6 & \underline{62.1} & 55.9 & 59.3 & \underline{66.7} & 70.5 & \textbf{91.9} & \textbf{53.0} & \underline{94.4} & \underline{80.0} & \underline{55.6} & \underline{94.7} & \underline{87.1} & 88.6 & \underline{82.1} & \underline{65.0} & \underline{68.8} & \underline{76.1} \\
Ours & \textbf{98.1} & \textbf{83.1} & \textbf{93.5} & \textbf{64.1}& \textbf{60.4} & \textbf{61.2} & \textbf{68.3} & \textbf{73.3} & 91.4 & \underline{52.1} & \textbf{95.0} & \textbf{80.3} & 55.2 & \textbf{94.8} & \textbf{87.7} & \textbf{91.8} & \textbf{83.4} & \textbf{66.1} & \textbf{70.0} & \textbf{77.4} \\
\toprule
\multicolumn{21}{c}{\textbf{Synthetic-to-Real: Synthia$\to$Cityscapes (Val.)}} \\
\hline

ADVENT~\cite{vu2019adventadversarialentropyminimization} & 85.6 & 42.2 & 79.7 & 8.7 & 0.4 & 25.9 & 5.4 & 8.1 & 80.4 & -- & 84.1 & 57.9 & 23.8 & 73.3 & -- & 36.4 & -- & 14.2 & 33.0 & 41.2\\
DACS~\cite{tranheden2021dacs} & 80.6 & 25.1 & 81.9 & 21.5 & 2.9 & 37.2 & 22.7 & 24.0 & 83.7 & -- & 90.8 & 67.6 & 38.3 & 82.9 & -- & 38.9 & -- & 28.5 & 47.6 & 48.3\\
ProDA~\cite{zhang2021prototypical} & \underline{87.8} & 45.7 & 84.6 & 37.1 & 0.6 & 44.0 & 54.6 & 37.0 & \textbf{88.1} & -- & 84.4 & 74.2 & 24.3 & 88.2 & -- & 51.1 & -- & 40.5 & 45.6 & 55.5\\
DAFormer~\cite{hoyer2022daformerimprovingnetworkarchitectures} & 84.5 & 40.7 & 88.4 & 41.5 & 6.5 & 50.0 & 55.0 & 54.6 & 86.0 & -- & 89.8 & 73.2 & 48.2 & 87.2 & -- & 53.2 & -- & 53.9 & 61.7 & 60.9\\
HRDA~\cite{hoyer2022hrdacontextawarehighresolutiondomainadaptive} & 85.2 & 47.7 & 88.8 & 49.5 & 4.8 & 57.2 & 65.7 & 60.9 & 85.3 & -- & 92.9 & 79.4 & 52.8 & 89.0 & -- & 64.7 & -- & 63.9 & 64.9 & 65.8\\
MIC~\cite{hoyer2023micmaskedimageconsistency} & 86.6 & \underline{50.5} & 89.3 & 47.9 & 7.8 & 59.4 & 66.7 & \underline{63.4} & 87.1 & -- & 94.6 & \underline{81.0} & \textbf{58.9} & \textbf{90.1} & -- & 61.9 & -- & 67.1 & 64.3 & 67.3\\
COPT~\cite{mata2025coptunsuperviseddomainadaptive} & 83.4 & 44.3 & \underline{90.0} & \underline{50.4} & \underline{8.0} & \underline{60.0} & \underline{67.0} & 63.0 & \underline{87.5} & -- & \underline{94.8} & \textbf{81.1} & \underline{58.6} & \underline{89.7} & -- & \underline{66.5} & -- & \underline{68.9} & \underline{65.0} & \underline{67.4} \\

Ours & \textbf{89.4} & \textbf{52.7} & \textbf{91.4} & \textbf{51.1} & \textbf{10.3} & \textbf{60.7} & \textbf{68.1} & \textbf{64.8} & 87.3 & -- & \textbf{95.1} & 80.6 & 57.4 & 89.2 & -- & \textbf{67.8} & -- & \textbf{69.3} & \textbf{65.4} & \textbf{68.8} \\

\hline

\bottomrule
\end{tabular}
\end{adjustbox}
\end{table}

\newcommand{\tsnelegend}{
\begin{tikzpicture}[font=\fontsize{8}{7}\selectfont]
\def\xbox{0}
\def\xtext{0.45}
\def\dy{0.25}

\foreach \name/\r/\g/\b/\i in {
road/128/64/128/0,
sidewalk/244/35/232/1,
building/70/70/70/2,
wall/102/102/156/3,
fence/190/153/153/4,
pole/153/153/153/5,
traffic light/250/170/30/6,
traffic sign/220/220/0/7,
vegetation/107/142/35/8,
terrain/152/251/152/9,
sky/70/130/180/10,
person/220/20/60/11,
rider/255/0/0/12,
car/0/0/142/13,
truck/0/0/70/14,
bus/0/60/100/15,
train/0/80/100/16,
motorcycle/0/0/230/17,
bicycle/119/11/32/18
}{
  \definecolor{legendcolor\i}{RGB}{\r,\g,\b}
  \filldraw[draw=black, fill=legendcolor\i] 
    (\xbox,-\i*\dy) rectangle ++(0.28,0.18);
  \node[anchor=west] at (\xtext,-\i*\dy+0.09) {\name};
}
\end{tikzpicture}
}

\begin{figure}[t]
\vspace{-0.4cm}
\centering

\begin{subfigure}[t]{0.33\textwidth}
\centering
\includegraphics[width=\linewidth]{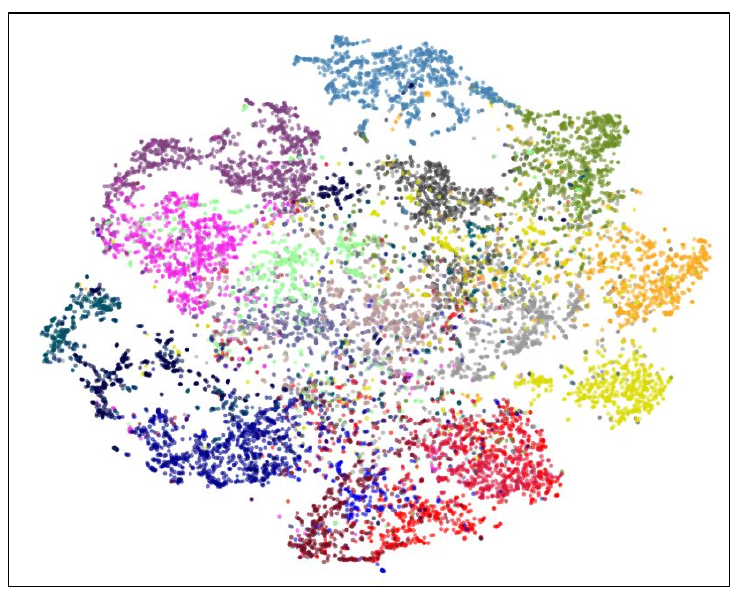}
\caption{Baseline}
\label{fig:tsne_baseline}
\end{subfigure}
\hspace{0.03\textwidth}%
\begin{subfigure}[t]{0.33\textwidth}
\centering
\includegraphics[width=\linewidth]{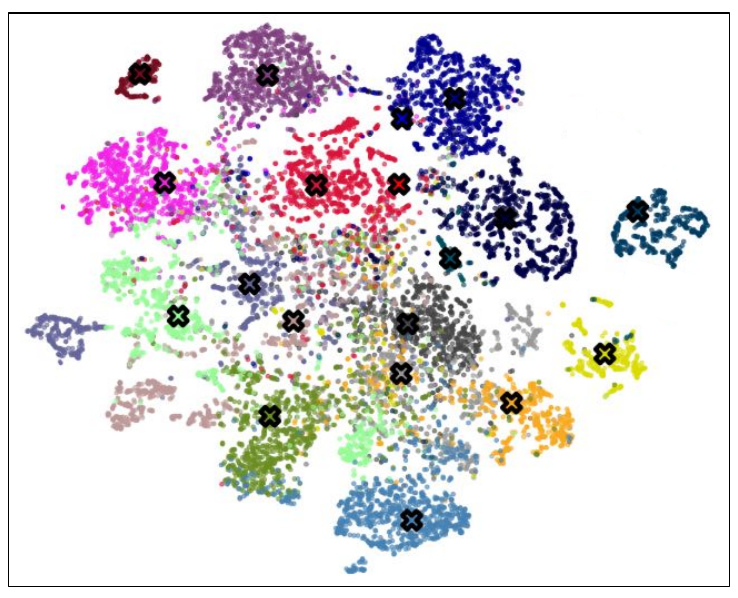}
\caption{Ours}
\label{fig:tsne_ours}
\end{subfigure}
\begin{subfigure}[t]{0.28\textwidth}
\centering
\raisebox{-0.55cm}{%
  \resizebox{0.58\linewidth}{!}{\tsnelegend}%
}
\label{fig:tsne_class_label}
\end{subfigure}

\caption{\textbf{t-SNE visualization on Cityscapes val set.}
Compared with the baseline without SAM and DINO, ours forms more compact target-feature clusters around DINOv3 prototype anchors (black crosses), indicating improved prototype-guided alignment.}
\label{fig:tsne}

\end{figure}
\vspace{-0.8cm}

\subsection{Ablation}
\vspace{-0.2cm}
\label{ablation}
Table~\ref{tab:other_uda} presents an ablation study evaluating the impact of our proposed components ---DINO prototype-based feature alignment and superpixel-guided SAM pseudo-label refinement---across different network architectures and UDA methods. Incorporating our components consistently improves performance across all evaluated backbones and UDA frameworks, demonstrating the general applicability of our approach.
We observe several interesting trends in the results. First, DINO prototype-based feature alignment yields larger performance gains for DeepLabV2 compared to DAFormer. This suggests that aligning CNN-based features with DINO prototypes, which encode rich semantic knowledge and strong generalization capabilities learned from large-scale data, effectively compensates for representation limitations inherent to CNN architectures. Second, SAM-based pseudo-label refinement becomes more beneficial as baseline performance increases. When the underlying UDA method produces more reliable pseudo-labels, SAM can refine a larger set of accurate regions, providing stronger and cleaner supervision to higher-performing baselines. Finally, we observe diminishing absolute performance gains when applying our method to stronger UDA baselines, a trend commonly observed in state-of-the-art adaptation methods due to performance saturation. Qualitative visualizations corresponding to these ablation results are provided in Fig.~\ref{fig2}.
\vspace{-0.3cm}

\begin{table}
\centering
\caption{Segmentation performance (mIoU in \%) of our method with different UDA methods on GTA$\to$CS.}
\label{tab:other_uda}
\setlength{\tabcolsep}{5pt}
\scriptsize
\begin{tabular}{llcccc}
\hline
Network & UDA Method & w/o SAM + DINO & w/ DINO & w/ SAM + DINO & Diff. \\
\hline\hline
DeepLabV2 \cite{chen2017deeplabsemanticimagesegmentation} &       DACS \cite{tranheden2021dacs} &    55.93 &   58.12 & 58.77 &        +2.84 \\
DeepLabV2 \cite{chen2017deeplabsemanticimagesegmentation} &   DAFormer \cite{hoyer2022daformerimprovingnetworkarchitectures} &    57.55 &   59.48 & 60.37 &        +2.82 \\
DeepLabV2 \cite{chen2017deeplabsemanticimagesegmentation} &           HRDA \cite{hoyer2022hrdacontextawarehighresolutiondomainadaptive} &    62.74 &   63.41 & 64.80 &      +2.06 \\
DeepLabV2 \cite{chen2017deeplabsemanticimagesegmentation} &           MIC \cite{hoyer2023micmaskedimageconsistency} &    63.53 &   64.81 & 65.36 &      +1.83 \\
\hline
DAFormer \cite{hoyer2022daformerimprovingnetworkarchitectures} &   DAFormer \cite{hoyer2022daformerimprovingnetworkarchitectures} &    66.72 &   68.45 & 69.17 &      +2.45 \\
DAFormer \cite{hoyer2022daformerimprovingnetworkarchitectures} &           HRDA \cite{hoyer2022hrdacontextawarehighresolutiondomainadaptive} &    73.72 &   74.57 &  76.04 &      +2.32 \\
DAFormer \cite{hoyer2022daformerimprovingnetworkarchitectures} &           MIC \cite{hoyer2023micmaskedimageconsistency} &    75.19 &   76.18 &  77.4 &      +2.21 \\
\hline
\end{tabular}
\end{table}
\vspace{-0.5cm}

%% file: conclusions.tex
\section{Conclusions}
\vspace{-0.2cm}
In this paper, we presented a dual-foundation framework for unsupervised domain adaptation in semantic segmentation that addresses two key limitations of existing methods: restricted supervision from high-confidence pseudo-labels and unstable, source-biased prototype representations. To this end, we leverage SAM with superpixel-based prompting and mask filtering to effectively refine object-level pseudo-labels, and DINOv3-based class prototypes to provide stable feature-alignment guidance, particularly for rare and challenging classes. Experiments on GTA$\rightarrow$Cityscapes and SYNTHIA$\rightarrow$Cityscapes show consistent improvements of +1.3\% and +1.4\% mIoU over strong baselines without replacing standard segmentation backbones. This work demonstrates that domain-robust foundation models can serve as effective auxiliary guidance, offering a scalable strategy for future domain adaptation research.
\vspace{-0.1cm}

%% file: acknowledgements.tex
\section*{Acknowledgments}
\vspace{-0.3cm}
This work was supported by the Institute of Information \& Communications Technology Planning \& Evaluation(IITP)-Innovative Human Resource Development for Local Intellectualization program grant funded by the Korea government(MSIT) (IITP-2025-RS-2023-00259678).
\vspace{-0.2cm}

%% file: references.bib
@article{toldo2020unsupervised,
  title={Unsupervised domain adaptation in semantic segmentation: a review},
  author={Toldo, Marco and Maracani, Andrea and Michieli, Umberto and Zanuttigh, Pietro},
  journal={Technologies},
  volume={8},
  number={2},
  pages={35},
  year={2020},
  publisher={MDPI}
}

@inproceedings{cordts2016cityscapes,
  title={The cityscapes dataset for semantic urban scene understanding},
  author={Cordts, Marius and Omran, Mohamed and Ramos, Sebastian and Rehfeld, Timo and Enzweiler, Markus and Benenson, Rodrigo and Franke, Uwe and Roth, Stefan and Schiele, Bernt},
  booktitle={CVPR},
  year={2016}
}

@inproceedings{sakaridis2021acdc,
  title={ACDC: The adverse conditions dataset with correspondences for semantic driving scene understanding},
  author={Sakaridis, Christos and Dai, Dengxin and Van Gool, Luc},
  booktitle={ICCV},
  year={2021}
}

@inproceedings{richter2016playing,
  title={Playing for data: Ground truth from computer games},
  author={Richter, Stephan R and Vineet, Vibhav and Roth, Stefan and Koltun, Vladlen},
  booktitle={ECCV},
  year={2016},
  organization={Springer}
}

@inproceedings{ros2016synthia,
  title={The synthia dataset: A large collection of synthetic images for semantic segmentation of urban scenes},
  author={Ros, German and Sellart, Laura and Materzynska, Joanna and Vazquez, David and Lopez, Antonio M},
  booktitle={CVPR},
  year={2016}
}

@inproceedings{gong2019dlow,
  title={Dlow: Domain flow for adaptation and generalization},
  author={Gong, Rui and Li, Wen and Chen, Yuhua and Gool, Luc Van},
  booktitle={CVPR},
  year={2019}
}

@inproceedings{yang2020fda,
  title={Fda: Fourier domain adaptation for semantic segmentation},
  author={Yang, Yanchao and Soatto, Stefano},
  booktitle={CVPR},
  year={2020}
}

@inproceedings{kim2020learning,
  title={Learning texture invariant representation for domain adaptation of semantic segmentation},
  author={Kim, Myeongjin and Byun, Hyeran},
  booktitle={CVPR},
  year={2020}
}

@inproceedings{melas2021pixmatch,
  title={Pixmatch: Unsupervised domain adaptation via pixelwise consistency training},
  author={Melas-Kyriazi, Luke and Manrai, Arjun K},
  booktitle={CVPR},
  year={2021}
}

@inproceedings{paul2020domain,
  title={Domain adaptive semantic segmentation using weak labels},
  author={Paul, Sujoy and Tsai, Yi-Hsuan and Schulter, Samuel and Roy-Chowdhury, Amit K and Chandraker, Manmohan},
  booktitle={ECCV},
  year={2020}
}

@inproceedings{das2023weakly,
  title={Weakly-supervised domain adaptive semantic segmentation with prototypical contrastive learning},
  author={Das, Anurag and Xian, Yongqin and Dai, Dengxin and Schiele, Bernt},
  booktitle={CVPR},
  year={2023}
}

@inproceedings{jiang2022prototypical,
  title={Prototypical contrast adaptation for domain adaptive semantic segmentation},
  author={Jiang, Zhengkai and Li, Yuxi and Yang, Ceyuan and Gao, Peng and Wang, Yabiao and Tai, Ying and Wang, Chengjie},
  booktitle={ECCV},
  year={2022},
}

@inproceedings{tranheden2021dacs,
  title={Dacs: Domain adaptation via cross-domain mixed sampling},
  author={Tranheden, Wilhelm and Olsson, Viktor and Pinto, Juliano and Svensson, Lennart},
  booktitle={WACV},
  year={2021}
}

@article{berthelot2019mixmatch,
  title={Mixmatch: A holistic approach to semi-supervised learning},
  author={Berthelot, David and Carlini, Nicholas and Goodfellow, Ian and Papernot, Nicolas and Oliver, Avital and Raffel, Colin A},
  journal={NeurIPS},
  year={2019}
}

@article{kang2020pixel,
  title={Pixel-level cycle association: A new perspective for domain adaptive semantic segmentation},
  author={Kang, Guoliang and Wei, Yunchao and Yang, Yi and Zhuang, Yueting and Hauptmann, Alexander},
  journal={NeurIPS},
  year={2020}
}

@inproceedings{wang2021exploring,
  title={Exploring cross-image pixel contrast for semantic segmentation},
  author={Wang, Wenguan and Zhou, Tianfei and Yu, Fisher and Dai, Jifeng and Konukoglu, Ender and Van Gool, Luc},
  booktitle={ICCV},
  year={2021}
}

@inproceedings{zhang2021prototypical,
  title={Prototypical pseudo label denoising and target structure learning for domain adaptive semantic segmentation},
  author={Zhang, Pan and Zhang, Bo and Zhang, Ting and Chen, Dong and Wang, Yong and Wen, Fang},
  booktitle={CVPR},
  year={2021}
}

@inproceedings{kirillov2023segment,
  title={Segment anything},
  author={Kirillov, Alexander and Mintun, Eric and Ravi, Nikhila and Mao, Hanzi and Rolland, Chloe and Gustafson, Laura and Xiao, Tete and Whitehead, Spencer and Berg, Alexander C and Lo, Wan-Yen and others},
  booktitle={ICCV},
  year={2023}
}

@article{simeoni2025dinov3,
  title={Dinov3},
  author={Sim{\'e}oni, Oriane and Vo, Huy V and Seitzer, Maximilian and Baldassarre, Federico and Oquab, Maxime and Jose, Cijo and Khalidov, Vasil and Szafraniec, Marc and Yi, Seungeun and Ramamonjisoa, Micha{\"e}l and others},
  journal={arXiv preprint arXiv:2508.10104},
  year={2025}
}

@inproceedings{bruggemann2023refignalignrefineadaptation,
  title={Refign: Align and refine for adaptation of semantic segmentation to adverse conditions},
  author={Br{\"u}ggemann, David and Sakaridis, Christos and Truong, Prune and Van Gool, Luc},
  booktitle={WACV},
  year={2023}
}

@inproceedings{hoyer2022daformerimprovingnetworkarchitectures,
  title={Daformer: Improving network architectures and training strategies for domain-adaptive semantic segmentation},
  author={Hoyer, Lukas and Dai, Dengxin and Van Gool, Luc},
  booktitle={CVPR},
  year={2022}
}

@inproceedings{wang2022continualtesttimedomainadaptation,
  title={Continual test-time domain adaptation},
  author={Wang, Qin and Fink, Olga and Van Gool, Luc and Dai, Dengxin},
  booktitle={CVPR},
  year={2022}
}

@article{tarvainen2018meanteachersbetterrole,
  title={Mean teachers are better role models: Weight-averaged consistency targets improve semi-supervised deep learning results},
  author={Tarvainen, Antti and Valpola, Harri},
  journal={NeuRIPS},
  year={2017}
}

@misc{yang2022divideadaptmitigatingconfirmation,
      title={Divide to Adapt: Mitigating Confirmation Bias for Domain Adaptation of Black-Box Predictors}, 
      author={Jianfei Yang and Xiangyu Peng and Kai Wang and Zheng Zhu and Jiashi Feng and Lihua Xie and Yang You},
      year={2022},
      eprint={2205.14467},
      archivePrefix={arXiv},
      primaryClass={cs.LG},
      url={https://arxiv.org/abs/2205.14467}, 
      note={Accessed 3 May 2026}
}

@inproceedings{arpit2017closerlookmemorizationdeep,
  title={A closer look at memorization in deep networks},
  author={Arpit, Devansh and Jastrz{\k{e}}bski, Stanis{\l}aw and Ballas, Nicolas and Krueger, David and Bengio, Emmanuel and Kanwal, Maxinder S and Maharaj, Tegan and Fischer, Asja and Courville, Aaron and Bengio, Yoshua and others},
  booktitle={ICML},
  year={2017}
}

@InProceedings{10.1007/978-3-030-58568-6_26,
author="Li, Guangrui
and Kang, Guoliang
and Liu, Wu
and Wei, Yunchao
and Yang, Yi",
editor="Vedaldi, Andrea
and Bischof, Horst
and Brox, Thomas
and Frahm, Jan-Michael",
title="Content-Consistent Matching for Domain Adaptive Semantic Segmentation",
booktitle="ECCV",
year="2020"
}

@InProceedings{Wang_2020_CVPR,
author = {Wang, Zhonghao and Yu, Mo and Wei, Yunchao and Feris, Rogerio and Xiong, Jinjun and Hwu, Wen-mei and Huang, Thomas S. and Shi, Honghui},
title = {Differential Treatment for Stuff and Things: A Simple Unsupervised Domain Adaptation Method for Semantic Segmentation},
booktitle = {CVPR},
year = {2020}
}

@inproceedings{subhani2020learning,
  title={Learning from scale-invariant examples for domain adaptation in semantic segmentation},
  author={Subhani, M Naseer and Ali, Mohsen},
  booktitle={ECCV},
  year={2020}
}

@inproceedings{wang2021uncertainty,
  title={Uncertainty-aware pseudo label refinery for domain adaptive semantic segmentation},
  author={Wang, Yuxi and Peng, Junran and Zhang, ZhaoXiang},
  booktitle={ICCV},
  year={2021}
}

@inproceedings{zhao2023unsuperviseddomainadaptationsemantic,
  title={Unsupervised domain adaptation for semantic segmentation with pseudo label self-refinement},
  author={Zhao, Xingchen and Mithun, Niluthpol Chowdhury and Rajvanshi, Abhinav and Chiu, Han-Pang and Samarasekera, Supun},
  booktitle={WACV},
  year={2024}
}

@inproceedings{guo2021metacorrectiondomainawaremetaloss,
  title={Metacorrection: Domain-aware meta loss correction for unsupervised domain adaptation in semantic segmentation},
  author={Guo, Xiaoqing and Yang, Chen and Li, Baopu and Yuan, Yixuan},
  booktitle={CVPR},
  year={2021}
}

@article{liu2025srplsfdasamguidedreliablepseudolabels,
  title={SRPL-SFDA: SAM-Guided Reliable Pseudo-Labels for Source-Free Domain Adaptation in Medical Image Segmentation},
  author={Liu, Xinya and Wu, Jianghao and Lu, Tao and Zhang, Shaoting and Wang, Guotai},
  journal={Neurocomputing},
  pages={130749},
  year={2025},
  publisher={Elsevier}
}

@inproceedings{qin2024langsplat3dlanguagegaussian,
  title={Langsplat: 3d language gaussian splatting},
  author={Qin, Minghan and Li, Wanhua and Zhou, Jiawei and Wang, Haoqian and Pfister, Hanspeter},
  booktitle={CVPR},
  year={2024}
}

@article{peng2023sam,
  title={Sam-guided unsupervised domain adaptation for 3d segmentation},
  author={Peng, Xidong and Chen, Runnan and Qiao, Feng and Kong, Lingdong and Liu, Youquan and Wang, Tai and Zhu, Xinge and Ma, Yuexin},
  year={2023}
}

@misc{lin2025samrefinertamingsegmentmodel,
      title={SAMRefiner: Taming Segment Anything Model for Universal Mask Refinement}, 
      author={Yuqi Lin and Hengjia Li and Wenqi Shao and Zheng Yang and Jun Zhao and Xiaofei He and Ping Luo and Kaipeng Zhang},
      year={2025},
      eprint={2502.06756},
      archivePrefix={arXiv},
      primaryClass={cs.CV}, 
      url={https://arxiv.org/abs/2502.06756}, 
      note={Accessed 3 May 2026}
}

@ARTICLE{sam4udass,
  author={Yan, Weihao and Qian, Yeqiang and Zhuang, Hanyang and Wang, Chunxiang and Yang, Ming},
  journal={Transactions on Intelligent Vehicles}, 
  title={SAM4UDASS: When SAM Meets Unsupervised Domain Adaptive Semantic Segmentation in Intelligent Vehicles}, 
  year={2024},
  volume={9},
  number={2},
  pages={3396-3408},
  doi={10.1109/TIV.2023.3344754},
  note={Accessed 3 May 2026}}

@inproceedings{benigmim2024collaboratingfoundationmodelsdomain,
  title={Collaborating foundation models for domain generalized semantic segmentation},
  author={Benigmim, Yasser and Roy, Subhankar and Essid, Slim and Kalogeiton, Vicky and Lathuili{\`e}re, St{\'e}phane},
  booktitle={CVPR},
  year={2024}
}

@article{hoshen1976percolation,
  title={Percolation and cluster distribution. I. Cluster multiple labeling technique and critical concentration algorithm},
  author={Hoshen, Joseph and Kopelman, Raoul},
  journal={Physical Review B},
  volume={14},
  number={8},
  pages={3438},
  year={1976},
  publisher={APS}
}

@inproceedings{englert2024exploringbenefitsvisionfoundation,
  title={Exploring the benefits of vision foundation models for unsupervised domain adaptation},
  author={Englert, Brun{\'o} B and Piva, Fabrizio J and Kerssies, Tommie and De Geus, Daan and Dubbelman, Gijs},
  booktitle={CVPR},
  year={2024}
}

@inproceedings{10.1145/3664647.3680582,
author = {Wu, Yao and Xing, Mingwei and Zhang, Yachao and Xie, Yuan and Qu, Yanyun},
title = {CLIP2UDA: Making Frozen CLIP Reward Unsupervised Domain Adaptation in 3D Semantic Segmentation},
year = {2024},
booktitle = {ACM Multimedia}
}

@inproceedings{fahes2023podapromptdrivenzeroshotdomain,
  title={Poda: Prompt-driven zero-shot domain adaptation},
  author={Fahes, Mohammad and Vu, Tuan-Hung and Bursuc, Andrei and P{\'e}rez, Patrick and De Charette, Raoul},
  booktitle={ICCV},
  year={2023}
}

@inproceedings{yang2024unifiedlanguagedrivenzeroshotdomain,
  title={Unified language-driven zero-shot domain adaptation},
  author={Yang, Senqiao and Tian, Zhuotao and Jiang, Li and Jia, Jiaya},
  booktitle={CVPR},
  year={2024}
}

@InProceedings{Sikdar_2025_CVPR,
    author    = {Sikdar, Aniruddh and Kishor, Arya and Kadam, Ishika and Sundaram, Suresh},
    title     = {PiCaZo: Pixel-Aligned Contrastive Learning for Zero-Shot Domain Adaptation},
    booktitle = {CVPR Workshops},
    year      = {2025}
}

@inproceedings{mata2025coptunsuperviseddomainadaptive,
  title={Copt: Unsupervised domain adaptive segmentation using domain-agnostic text embeddings},
  author={Mata, Cristina and Ranasinghe, Kanchana and Ryoo, Michael S},
  booktitle={ECCV},
  year={2024}
}

@misc{liu2025langdabuildingcontextawarenesslanguage,
      title={LangDA: Building Context-Awareness via Language for Domain Adaptive Semantic Segmentation}, 
      author={Chang Liu and Bavesh Balaji and Saad Hossain and C Thomas and Kwei-Herng Lai and Raviteja Vemulapalli and Alexander Wong and Sirisha Rambhatla},
      year={2025},
      eprint={2503.12780},
      archivePrefix={arXiv},
      primaryClass={cs.CV},
      url={https://arxiv.org/abs/2503.12780}, 
      note={Accessed 3 May 2026}
}

@misc{abedi2024eudaefficientunsuperviseddomain,
      title={EUDA: An Efficient Unsupervised Domain Adaptation via Self-Supervised Vision Transformer}, 
      author={Ali Abedi and Q. M. Jonathan Wu and Ning Zhang and Farhad Pourpanah},
      year={2024},
      eprint={2407.21311},
      archivePrefix={arXiv},
      primaryClass={cs.CV},
      url={https://arxiv.org/abs/2407.21311}, 
      note={Accessed 3 May 2026}
      
}

@inproceedings{fahes2024simplerecipelanguageguideddomain,
  title={A simple recipe for language-guided domain generalized segmentation},
  author={Fahes, Mohammad and Vu, Tuan-Hung and Bursuc, Andrei and P{\'e}rez, Patrick and De Charette, Raoul},
  booktitle={CVPR},
  year={2024}
}

@inproceedings{radford2021learningtransferablevisualmodels,
  title={Learning transferable visual models from natural language supervision},
  author={Radford, Alec and Kim, Jong Wook and Hallacy, Chris and Ramesh, Aditya and Goh, Gabriel and Agarwal, Sandhini and Sastry, Girish and Askell, Amanda and Mishkin, Pamela and Clark, Jack and others},
  booktitle={ICML},
  year={2021}
}

@misc{oquab2024dinov2learningrobustvisual,
      title={DINOv2: Learning Robust Visual Features without Supervision}, 
      author={Maxime Oquab and Timothée Darcet and Théo Moutakanni and Huy Vo and Marc Szafraniec and Vasil Khalidov and Pierre Fernandez and Daniel Haziza and Francisco Massa and Alaaeldin El-Nouby and Mahmoud Assran and Nicolas Ballas and Wojciech Galuba and Russell Howes and Po-Yao Huang and Shang-Wen Li and Ishan Misra and Michael Rabbat and Vasu Sharma and Gabriel Synnaeve and Hu Xu and Hervé Jegou and Julien Mairal and Patrick Labatut and Armand Joulin and Piotr Bojanowski},
      year={2024},
      eprint={2304.07193},
      archivePrefix={arXiv},
      primaryClass={cs.CV},
      url={https://arxiv.org/abs/2304.07193}, 
      note={Accessed 3 May 2026}
}

@inproceedings{hoyer2022hrdacontextawarehighresolutiondomainadaptive,
  title={HRDA: Context-aware high-resolution domain-adaptive semantic segmentation},
  author={Hoyer, Lukas and Dai, Dengxin and Van Gool, Luc},
  booktitle={ECCV},
  year={2022}
}

@inproceedings{hoyer2023micmaskedimageconsistency,
  title={MIC: Masked image consistency for context-enhanced domain adaptation},
  author={Hoyer, Lukas and Dai, Dengxin and Wang, Haoran and Van Gool, Luc},
  booktitle={CVPR},
  year={2023}
}

@article{chen2017deeplabsemanticimagesegmentation,
  title={Deeplab: Semantic image segmentation with deep convolutional nets, atrous convolution, and fully connected crfs},
  author={Chen, Liang-Chieh and Papandreou, George and Kokkinos, Iasonas and Murphy, Kevin and Yuille, Alan L},
  journal={T-PAMI},
  volume={40},
  number={4},
  pages={834--848},
  year={2017},
  publisher={IEEE}
}

@inproceedings{vu2019adventadversarialentropyminimization,
  title={Advent: Adversarial entropy minimization for domain adaptation in semantic segmentation},
  author={Vu, Tuan-Hung and Jain, Himalaya and Bucher, Maxime and Cord, Matthieu and P{\'e}rez, Patrick},
  booktitle={CVPR},
  year={2019}
}

@article{badrinarayanan2016segnetdeepconvolutionalencoderdecoder,
  title={Segnet: A deep convolutional encoder-decoder architecture for image segmentation},
  author={Badrinarayanan, Vijay and Kendall, Alex and Cipolla, Roberto},
  journal={T-PAMI},
  volume={39},
  number={12},
  pages={2481--2495},
  year={2017},
  publisher={IEEE}
}

@inproceedings{mccormac2016semanticfusiondense3dsemantic,
  title={Semanticfusion: Dense 3d semantic mapping with convolutional neural networks},
  author={McCormac, John and Handa, Ankur and Davison, Andrew and Leutenegger, Stefan},
  booktitle={ICRA},
  year={2017}
}

@InProceedings{Kweon_2024_CVPR,
    author    = {Kweon, Hyeokjun and Kim, Jihun and Yoon, Kuk-Jin},
    title     = {Weakly Supervised Point Cloud Semantic Segmentation via Artificial Oracle},
    booktitle = {CVPR)},
    year      = {2024}
}

@inproceedings{bergh2013seedssuperpixelsextractedenergydriven,
  title={Seeds: Superpixels extracted via energy-driven sampling},
  author={Van den Bergh, Michael and Boix, Xavier and Roig, Gemma and De Capitani, Benjamin and Van Gool, Luc},
  booktitle={ECCV},
  year={2012}
}
